%% file: main.tex
\documentclass[10pt,twocolumn,letterpaper]{article}

\usepackage{arxiv}
\input{math_commands.tex}

\title{Continual Tool-Space Extension Learning for Multimodal Agents}

\author{
\makebox[\textwidth][c]{%
    \normalsize
    Jinbae Seo$^{1}$ \quad
    Jungin Park$^{2}$ \quad
    Hyunsung Jang$^{1,3}$ \quad
    Namkoo Ha$^{3}$ \quad
    Jiyoung Lee$^{4\dagger}$ \quad
    Kwanghoon Sohn$^{1\dagger}$
}
\\[0.8em]
\parbox{\textwidth}{%
    \centering\small
    $^{1}$Yonsei University \qquad
    $^{2}$DGIST \qquad
    $^{3}$LIG Defense \& Aerospace \\
    $^{4}$Division of Artificial Intelligence \& Software, Ewha Womans University \qquad
}
}

\begin{document}

\twocolumn[{%
\renewcommand\twocolumn[1][]{#1}%
\maketitle

\vspace{0em}

\begin{center}
    \centering
    \vspace{-1.5em}
    \includegraphics[width=\linewidth]{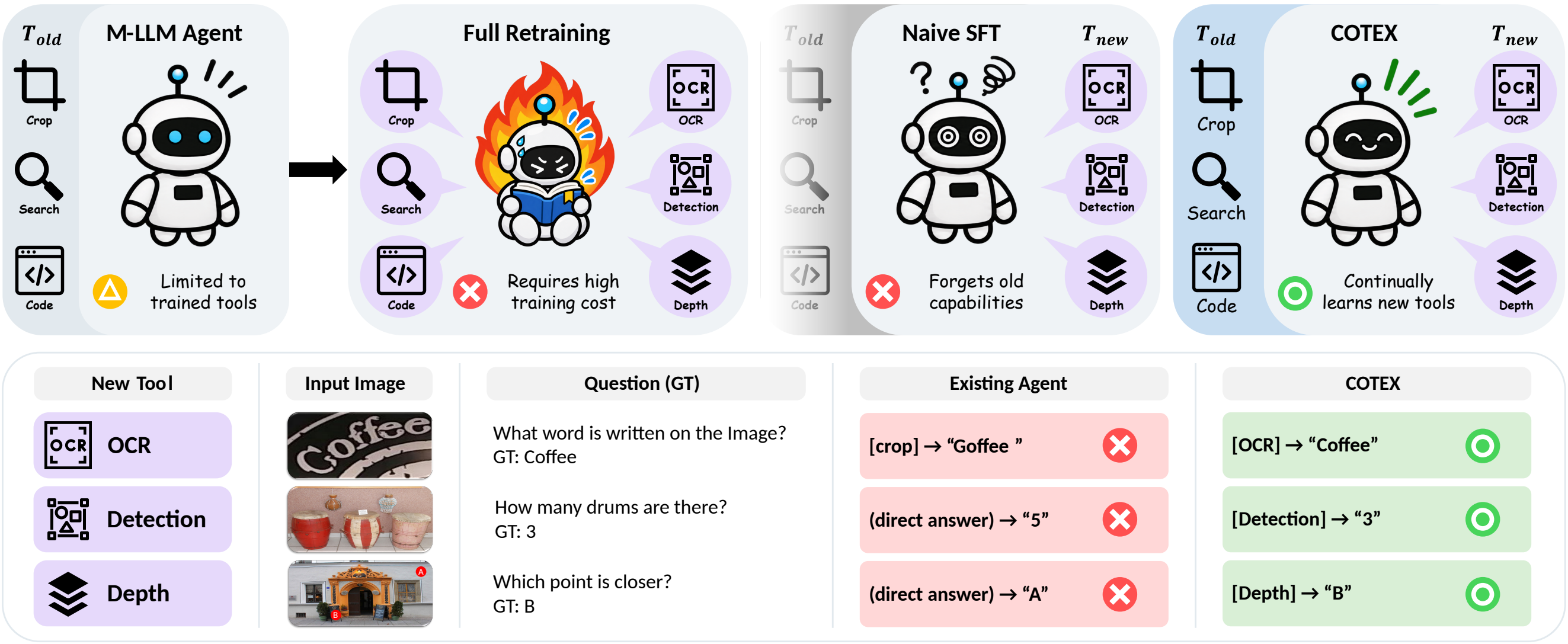}
    \captionof{figure}{
        Intuition of \ours: (a) full re-training preserves previously acquired tools but incurs high training cost, (b) naive parameter-efficient finetuning (PEFT)~\cite{hu2022lora} is efficient but overwrites previously acquired tools, (c) \ours, that weaves the advantages of both full re-training and PEFT techniques by orthogonally learning low-rank adapters for each novel tool.
    }
    \label{fig:teaser}
    \par\vspace{0.5em}
\end{center}%
}]

\begingroup
\renewcommand{\thefootnote}{}
\footnotetext[0]{$^{\dagger}$Corresponding authors}
\endgroup

\input{Sections/0.Abstract}
\input{Sections/1.Introduction}
\input{Sections/2.Related_Works}

\input{Sections/3.Method}

\input{Sections/4.Experiments}
\input{Sections/5.Conclusion}

\section*{LLM Usage Statement}
We used OpenAI ChatGPT to assist with experimental code development, LaTeX formatting, and language editing of the manuscript.
All AI-assisted code and manuscript content were reviewed and verified by the authors.
We take full responsibility for the final content of this work, including all experimental results, text, and artifacts.

{\small\input{main.bbl}}

\clearpage
\onecolumn
\appendix

\counterwithin{table}{section}
\counterwithin{figure}{section}
\counterwithin{algorithm}{section}
\counterwithin{equation}{section}

\begin{center}
    {\LARGE\bfseries
    Continual Tool-Space Extension Learning for Multimodal Agents
    \par}
    \vspace{0.5em}
    {\Large\bfseries -- Appendix --\par}
    \vspace{1.5em}
\end{center}

\input{Supple/A.Training}

\input{Supple/B.Exp}
\input{Supple/C.Limitation}
\input{Supple/prompt}


\clearpage

\end{document}

%% file: math_commands.tex
\usepackage{amsmath,amsfonts,bm}

\def\eqref#1{equation~\ref{#1}}

\def\1{\bm{1}}

\DeclareMathAlphabet{\mathsfit}{\encodingdefault}{\sfdefault}{m}{sl}
\SetMathAlphabet{\mathsfit}{bold}{\encodingdefault}{\sfdefault}{bx}{n}



%% file: Sections/0.Abstract.tex
\begin{abstract}
Multimodal agents largely rely on generic tools such as cropping, search, and code execution, leaving specialist vision models outside their tool space.
However, bringing such a model into an already trained agent is not straightforward.
The challenge is to learn each new tool efficiently while retaining the ability to use existing ones.
This work proposes \ours, continual tool-space extension learning, that adds new tools at efficient adapter-level cost.
\ours locates two sets of parameter directions; 
One triggers a tool call and thereby opens a multi-turn reasoning trajectory, while the other determines which tool is chosen.
The adapter of a new tool is restricted to the calling directions that leave earlier tool choices unaffected by orthogonal constraint.
We further introduce a protocol that converts specialist models' existing task-specific annotations into verified tool-use trajectories.
Experiments on two representative agents and three specialist tools show gains on nine vision-language benchmarks, with at most a 0.3-point drop on existing tools.
\end{abstract}

%% file: Sections/1.Introduction.tex
\section{Introduction}
\label{sec:introduction}
Multimodal large language models (MLLMs) have rapidly evolved from passive visual understanding systems into agentic models capable of interacting with external environments.
By integrating systematic operations into their reasoning process (\eg, code execution, search, and image cropping), recent multimodal agents such as DeepEyes~\cite{zheng2026deepeyes, hong2026deepeyesv2} and  Metis~\cite{yan2026metis} can actively acquire missing evidence beyond parametric inference alone.
Existing works have primarily focused on teaching agents how to invoke and utilize external tools.

In practice, however, the tool set is rarely fixed.
Many vision tasks are best solved by specialist models, and stronger specialists continue to appear.
As illustrated in \cref{fig:teaser}, an agent limited to cropping, search, and code often misreads scene text, miscounts objects, and misjudges relative depth.
Incorporating such tools (\eg, OCR model, a detector, or a depth estimator) after the agent has already been trained, which we refer to as \textbf{continual tool-space extension learning}, is therefore a natural requirement for a scalable agent.

Integrating newly available tools into an already trained agent introduces a new challenge.
Existing options sit at two ends of a trade-off between cost and retention.
A straightforward approach is agent full training, which re-runs SFT and RL over the expanded tool space, on newly induced trajectories with a novel tool.
Its cost, however, grows with every tool that is added.
Parameter-efficient fine-tuning (PEFT)~\cite{hu2022lora} avoids this cost, yet an adapter trained only on the new tool overwrites the directions the agent relies on for existing tools (\ie, catastrophic forgetting problem)~\cite{huang2024practical,liu2025colt}.
Consequently, a practical agent requires both efficient updates and intact existing tools.

In this paper, we propose a \textbf{CO}ntinual \textbf{T}ool-space \textbf{EX}tension learning (\textbf{\ours}).
\ours is built on the premise that tool use involves two distinct decisions.
Initiating a tool call reflects the judgment that the evidence at hand is insufficient to answer, which is common to every tool.
Selecting a tool determines what evidence is needed next, which is specific to each tool.
A new tool should therefore reuse the former and leave the latter untouched.
\ours applies SVD to the weight displacement between the agent and its base MLLM, then ranks the resulting directions by their gradient sensitivity on a small calibration set.
Directions sensitive to call initiation form the shared tool-call subspace.
Within it, directions sensitive to the selection of existing tools are marked as protected.
Each new tool is learned as a low-rank adapter that lies inside the shared subspace while having orthogonal directions to the protected directions.
The adapter thus inherits the judgment of when evidence is insufficient, while the choice among existing tools is left intact.

Learning the adapter still requires tool-use trajectories, which are rarely available for a newly released model.
We therefore introduce a general protocol that repurposes existing task-specific supervision into such trajectories.
Each tool corresponds to a well-defined task for which annotated data already exist, such as bounding boxes for a detector or depth maps for a depth estimator.
These annotations specify what a correct tool output should be, but they do not describe how an agent should invoke the tool and use its output to answer a question.
Our protocol bridges this gap with self-generated trajectories.
The agent is prompted to invoke the tool on examples from its native dataset, and a task-specific verifier retains only the trajectories whose final predictions agree with the original annotations.
Through training on these trajectories, the agent learns to decide whether to use the tool and how to invoke it on its own.

Extensive experiments demonstrate that \ours effectively grows two agents with three tools, including object detector~\cite{liu2024groundingdino}, OCR~\cite{Du2025SVTRv2}, and depth estimator~\cite{yang2024depthanything}.
With only about 1K self-generated demonstrations per tool, \ours improves performance across benchmarks requiring OCR, object detection, and depth estimation, while largely preserving the original agent's visual perception and reasoning capabilities.

In summary, this work provides key contributions:
\begin{itemize}
    \item We formulate continual tool-space extension, where a multimodal agent acquires new specialist tools while retaining existing ones.
    \item We propose \ours, which identifies the protected tool-sensitive parameters and learns each new tool orthogonally to it within the shared tool-call subspace.
    \item We introduce a training protocol that converts native task annotations into verified tool-use trajectories, and demonstrate across 10 benchmarks that \ours acquires specialized capabilities through parameter-efficient updates while preserving general visual perception performance.
\end{itemize}\vspace{-1em}

%% file: Sections/2.Related_Works.tex
\section{Related Work}
\label{sec:related_work}

\begin{figure*}[t]
    \centering
    \includegraphics[width=\linewidth]{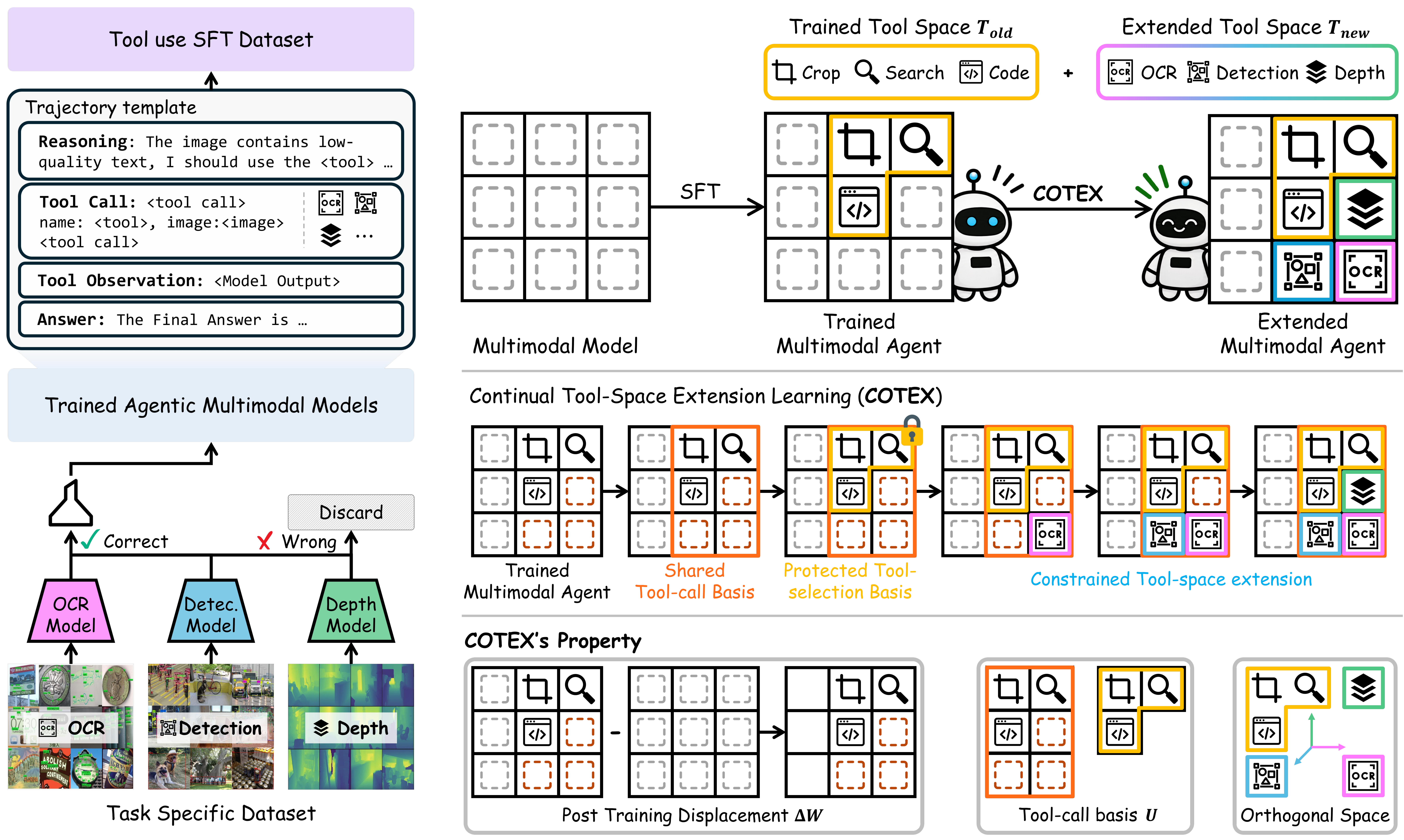}
    \vspace{-2em}
    \caption{The overall training pipeline of \ours.}
    \label{fig:fig3}
    \vspace{-0.5em}
\end{figure*}

\subsection{Multimodal Agents}
In general, multimodal agents~\cite{gupta2023visprog,zheng2026deepeyes} have built on MLLMs by explicitly executing image processing, computation, and information retrieval programs into the reasoning process.
Early models~\cite{gupta2023visprog,suris2023vipergpt} have generated executable program code in scripts, while GPT4Tools~\cite{yang2023gpt4tools} and LLaVA-Plus ~\cite{liu2024llavaplus} have learned to invoke predefined tools through instruction tuning within their reasoning chain.
Recently, reinforcement learning (RL)-based optimization~\cite{shao2024deepseekmath} has improved the scalability of tool-use in agents~\cite{zheng2026deepeyes,hong2026deepeyesv2,yan2026metis}.
For instance, DeepEyesV2~\cite{hong2026deepeyesv2} integrates visual operations, code execution, and external search into a unified reasoning policy.
Meanwhile, Metis~\cite{yan2026metis} addresses tool overuse by decoupling task correctness from interaction efficiency and discouraging redundant tool calls among successful trajectories.
These approaches substantially advance the acquisition and utilization of external tools, but mainly focus on improving tool use within a given interaction setting.
Our work instead considers how the available tool space itself can be expanded after post-training while preserving previously acquired tool-use behavior.

\subsection{Tool-Space Extension}
Tool invocation itself is learned quickly, yet sequential fine-tuning on a new tool erases previously learned ones almost entirely~\cite{huang2024practical}.
GeoPilot~\cite{zhou2026geopilot} repurposes existing task annotations to construct tool-use supervision, but trains with a predefined set of tools.
Even though COLT~\cite{liu2025colt} introduces a learnable tool codebook for single-turn API selection of Video-LLMs, it still finetunes the full model for each new tool, without controlling interference in weight space.
Instead, we target agentic visual reasoning, in which the forgetting problem is mitigated by restricting new tool updates to directions within the shared tool-call subspace that are orthogonal to previously learned tool-selection directions.

\subsection{Subspace Isolation and Adaptation}
Constraining parameter updates to a subspace adds new capabilities while limiting interference with existing knowledge~\cite{hu2022lora,dettmers2023qlora}.
Recently, Iso-CTS~\cite{marczak2025notask} identifies common and task-specific subspaces from weight differences for model merging, while DiFT~\cite{yuan2026dift} identifies capability-sensitive parameters from activation changes for selective fine-tuning.
Meanwhile, O-LoRA~\cite{wang2023olora} and KeepLoRA~\cite{luo2026keeplora} have constrained low-rank updates to be orthogonal to previously acquired knowledge, while  LoDA~\cite{he2026loda} has separated shared and task-specific subspaces to balance knowledge transfer and isolation.
Unlike prior orthogonal adaptation methods~\cite{wang2023olora,luo2026keeplora}, which steer new updates away from the dominant directions of the model, \ours places them inside the dominant tool-call directions and excludes only what governs existing tool selection.

%% file: Sections/3.Method.tex
\vspace{-0.5em}
\section{Method}
\label{sec:method}

\subsection{Problem Statement and Overview}
Let $\pi_\theta$ denote a multimodal agent obtained by post-training a base MLLM with parameters $\theta^\star$ on a tool set $\mathcal{T}_{\mathrm{old}}$ (\eg, cropping, Python execution, and search).
New tools $\{g_1, \dots, g_K \}$ arrive sequentially, where each $g_k$ is a frozen specialist model accessed through an executable interface.
When $g_k$ arrives, the goal is to enable $\pi_\theta$ to invoke it while retaining the use of $\mathcal{T}_{\mathrm{old}}$ and of the previously added tools $\{g_1, \dots, g_{k-1}\}$.
For efficient training, each tool is acquired through a dedicated low-rank update.

As illustrated in \cref{fig:fig3}, \ours consists of three components.
First, existing supervision for the task performed by $g_k$ is converted into a set of verified tool-use trajectories (\cref{sec:expandable_tools}).
Second, we locate the directions that drive call initiation, which might be updateable, and those that determine tool selection, which might be protected (\cref{sec:tool_call_analysis}).
Third, new tools are learned via a low-rank adapter inside the shared tool-call subspace while remaining orthogonal to protected tool-selection directions (\cref{sec:orthogonal_extension}).

\subsection{Task-Specific Models as Expandable Tools}
\label{sec:expandable_tools}
A robust model $g_k$ tailored for a specific computer vision task (\eg, object detection, OCR, and monocular depth estimation) could serve as a novel tool to employ an external capability on a multimodal agent.
Integrating a new tool $g_k$ requires trajectories that show when to call it, how to form its arguments, and how to use its observation.
Prior work synthesizes such trajectories with strong teacher model (\eg, GPT~\cite{hong2026deepeyesv2} or Gemini~\cite{wang2026beacon}), which is costly and ties data quality to the teacher.
We observe that the annotated dataset of $g_k$ already supplies the two ingredients a teacher would otherwise provide.
Its inputs are precisely the cases in which $g_k$ is useful.
Its ground-truths tell whether a final answer is correct.
With both in hand, trajectories need not be written by a teacher.
The agent attempts each example with access to $g_k$, and only the attempts whose final answer matches the ground-truths are kept.

Concretely, for an image input $x_i$, the agent emits a call $c_i$, receives the observation $o_i$ from executing $g_k$, and produces a prediction $\hat{y}_i$.
A task-specific verifier $\verifier_k$, implemented with rule-based criteria or an external judge, compares a prediction $\hat{y}_i$ with $y_i$.
Only trajectories $\tau_i$ verified as correct are retained in a task dataset,
\begin{equation}
    \mathcal{D}_k
    =
    \left\{
        (x_i,\tau_i)
        \,\middle|\,
        \mathcal{V}_k(\hat{y}_i,y_i)=1
    \right\},
    \label{eq:verified_tool_data}
\end{equation}
where $\mathcal{V}_k$ returns a binary verification result.
The instruction templates and verifier configurations for each task are provided in Appendix~A.3.
This construction allows existing task-specific supervision to serve directly as training data for tool integration.

\subsection{Low-Rank Structure of Tool-Call Sensitivity}
\label{sec:tool_call_analysis}

Introducing new tools requires adapting an existing agent without disrupting its previously acquired capabilities.
However, unconstrained parameter updates may disable tool-use, reasoning, and perception capabilities.
A natural remedy is to identify a shared subspace associated with tool-call behavior and protect the directions within it that are responsible for selecting existing tools.
We therefore investigate whether tool-call behavior is associated with a compact subspace of the parameter changes introduced during agent training.

For each adapted linear projection $W^\ell$ in the LLM, excluding vision-specific modules and the language head, we compute singular value decomposition (SVD) for weight displacement,
\begin{equation}
    \Delta W^\ell
    =
    W^\ell_{\theta} - W^\ell_{\theta^{\star}}
    =
    U_\Delta^\ell\Sigma^\ell{V^\ell}^\top,
    \label{eq:checkpoint_delta}
\end{equation}
where $U_\Delta^\ell \in \mathbb{R}^{d_{\mathrm{out}}\times d_\Delta}$ contains the left singular vectors of the post-training displacement, with $d_\Delta=\min(d_{\mathrm{out}},d_{\mathrm{in}})$.
These directions reflect every change introduced by post-training, including general reasoning, so not all of them concern tool use.
The sensitivity of tool-call behavior to each direction is therefore measured separately.

For a calibration trajectory $i$, let $\mathcal{I}_i$ denote the set of target token positions at which the loss is evaluated.
The calibration loss $\mathcal{L}_i$ is the mean negative log-likelihood of the target tokens in $\mathcal{I}_i$ under $\pi_\theta$, conditioned on the preceding interaction history:
\begin{equation}
    \mathcal{L}_i
    =
    -\frac{1}{|\mathcal{I}_i|}
    \sum_{t\in\mathcal{I}_i}
    \log \pi_{\theta}(y_{t}\mid h_{<t}),
\end{equation}
where $y_t$ is the token at position $t$ and $h_{<t}$ is the preceding interaction history, including tool observations.
A trajectory often contains more than one tool call.
For instance, an agent may crop a region, inspect it, and then call a numeric calculator.
Therefore, the trajectory has two positions where a call is initiated and two positions where a tool is named.
In other words, $\losscall$ is evaluated at positions that initiate a tool call (\eg, \tags{<code>}, \tags{<tool\_call>}), capturing when the agent decides to invoke a tool.
$\losstool$ is evaluated at positions that specify the selected tool (\eg, \tags{crop}, \tags{search}, \tags{image\_search}), capturing which tool the agent chooses.

\begin{figure}[t]
    \centering
    \includegraphics[width=0.8\linewidth]{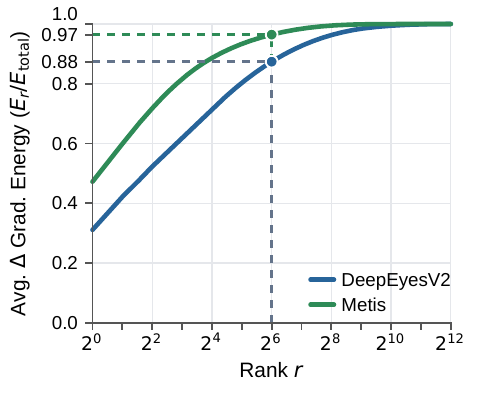}
    \vspace{-1.2em}
    \caption{Low-rank retention rate of tool-call gradient energy.
    Gradients of the tool-call loss are projected onto the top-$r$ tool-sensitive directions identified by $C(U_\Delta^\ell,\losscall)$ within the left singular-vector subspace of $\Delta W^\ell$, and the retained energy ratio $E_r/E_{\mathrm{total}}$ is averaged over layers.
    At $r{=}64$, under 2\% of the full rank, the projection retains 88\% of the energy for DeepEyesV2 and 97\% for Metis.
    }
    \label{fig:energy}
\end{figure}

Gradient-based tool sensitivity within an orthonormal basis $U$ is defined as
\begin{equation}
    C(U,\mathcal{L})
    =
    \frac{1}{N}
    \sum_{i=1}^{N}
    \left(U^\top \frac{\partial \mathcal{L}_i}{\partial W^\ell}\right)
    \left(U^\top \frac{\partial \mathcal{L}_i}{\partial W^\ell}\right)^\top,
    \label{eq:subspace_sensitivity}
\end{equation}
where $N$ is the number of calibration trajectories.
\cref{fig:energy} shows the cumulative gradient energy captured by the leading eigenvectors of $C(U_\Delta^\ell,\losscall)$, averaged across layers.
The result supports our hypothesis that tool-call sensitivity is concentrated in a low-rank subspace across different agents~\cite{hong2026deepeyesv2, yan2026metis}.
We next explain how to learn new tools within this shared subspace while remaining orthogonal to directions associated with existing tool selection.

\subsection{Orthogonal Tool Space Update}
\label{sec:orthogonal_extension}
To prevent interference between new tool usability and old-tool application capability, we constrain new tool updates to their remaining orthogonal subspace~\cite{wang2023olora, luo2026keeplora}.
Specifically, \ours's design departs from subspace-constrained adaptation such as KeepLoRA~\cite{luo2026keeplora} in two respects.
KeepLoRA~\cite{luo2026keeplora} decomposes the pre-trained weights and writes new knowledge into the residual subspace, away from their principal components.
\ours decomposes the post-training displacement rather than the weights themselves, ranks directions by their gradient sensitivity to tool-call behavior rather than by singular value magnitude, and places new tools inside the shared tool-call subspace.
Within this subspace, directions associated with existing tool selection are excluded from new-tool adaptation.
For notational simplicity, we omit the layer index $\ell$ in the following.

\paragrapht{Shared tool-call basis.}
Specifically, given a symmetric matrix $C$ by \cref{eq:subspace_sensitivity}, let $Q_{[j]}(C)$ denote the eigenvectors associated with its $j$ largest eigenvalues.
We define $d_{\mathrm{call}}$ as the dimensionality of the tool-call-sensitive subspace identified by $C(U_\Delta,\losscall)$.
We construct an orthonormal basis for the shared tool-call directions as
\begin{equation}
    \ucall
    =
    U_\Delta
    Q_{[d_{\mathrm{call}}]}
    \left(C(U_\Delta,\losscall)\right).
    \label{eq:shared_tool_space}
\end{equation}
The columns of $\ucall$ form an orthonormal basis within the dominant displacement subspace.

\paragrapht{Protected tool-selection basis.}
Within this shared subspace $\ucall$, we further identify directions sensitive to selecting previously learned tools,
\begin{equation}
    U_{\mathrm{old}}
    =
    U_{\mathrm{call}}
    \eigenvector
    \left(C(U_{\mathrm{call}},\losstool)\right),
    \label{eq:protected_tool_space}
\end{equation}
This hierarchical construction selects the protected directions within the shared tool-call subspace, leaving the remaining directions available for new-tool adaptation.

\paragrapht{Constrained new-tool adaptation.}
For a newly introduced tool $g_k$, let $U_{<k}$ be the accumulated orthonormal basis of protected tool-selection directions, initialized as $U_{\mathrm{old}}$.
Since these directions lie within the shared tool-call subspace, the projector onto the available adaptation space is
\begin{equation}
    P_k
    =
    U_{\mathrm{call}}U_{\mathrm{call}}^\top
    -
    U_{<k}U_{<k}^\top.
    \label{eq:available_tool_space}
\end{equation}
Let $R_{<k}$ be an orthonormal basis for the range of $P_k$, satisfying
$R_{<k}R_{<k}^\top=P_k$.
We employ a new tool's low-rank adapter~\cite{hu2022lora} as
\begin{equation}
    \Delta W_k
    =
    s_k R_{<k} B_k A_k,
    \label{eq:orthogonal_constraint}
\end{equation}
where $s_k$ is the scaling factor, and the fixed basis $R_{<k}$ confines every update to the shared tool-call subspace and excludes the accumulated protected directions.
$\Delta W_k$ has to be orthogonal to the accumulated protected tool-selection directions, satisfying $U_{<k}^{\top}\Delta W_k=0$.
Only $B_k$ and $A_k$ are optimized with the verified tool-use trajectories $\mathcal{D}_k$ described in \cref{sec:expandable_tools}.
Note that tool observations are included in the interaction history but excluded from the prediction targets.

\subsection{Training for Tool Utilization}
\label{sec:tool_rl}
\paragrapht{SFT.}
For each newly introduced tool $g_k$, the agent learns on the verified tool-use trajectories $\mathcal{D}_k$ by SFT.
We minimize the negative log-likelihood of the assistant-generated tokens, including reasoning, tool invocations, and final answers, conditioned on the preceding interaction history.
Tool observations are included in the history but excluded from the prediction targets.
During SFT, we freeze previously learned adapters, and the agent backbone.
The resulting updates are therefore confined to the available tool-call space and remain orthogonal to the accumulated protected tool-selection directions.

\paragrapht{RL.}
After SFT establishes basic tool-use capabilities, we further optimize the agent through reinforcement learning to improve when and how to utilize newly introduced tools.
Following Metis~\cite{yan2026metis}, we adopt Hierarchical Decoupled Policy Optimization (HDPO), which separately computes advantages for task outcomes and tool-use efficiency.
The outcome reward combines task accuracy and format compliance with weights $w_{\mathrm{acc}}$ and $w_{\mathrm{fmt}}$, while the efficiency reward favors fewer tool interactions among correct rollouts.
The resulting advantages are optimized through
\begin{equation}
    \mathcal{L}_{\mathrm{RL}}
    =
    \mathcal{L}_{\mathrm{GRPO}}(\textit{Adv}_{\mathrm{out}})
    +
    w_{\mathrm{eff}}\mathcal{L}_{\mathrm{GRPO}}(\textit{Adv}_{\mathrm{eff}}),
    \label{eq:rl_objective}
\end{equation}
where $\textit{Adv}_{\mathrm{out}}$ and $\textit{Adv}_{\mathrm{eff}}$ denote the independently computed outcome and efficiency advantages, respectively, and $w_{\mathrm{eff}}$ controls the contribution of tool-use efficiency.
During optimization, we update only the task-specific LoRA adapter while keeping the remaining adapters and backbone frozen.

\paragrapht{Continual protection.}
After learning $A_k$ and $B_k$, we evaluate its tool-selection calibration loss $\losstoolk$ on the adapted agent and extract the corresponding sensitive directions within the available adaptation space:
\begin{equation}
    U_k
    =
    R_{<k}
    \eigenvector
    \left(C(R_{<k},\losstoolk )\right).
    \label{eq:new_tool_subspace}
\end{equation}
We append $U_k$ to the accumulated protected basis $U_{<k}$ and use the updated basis $U_{\leq k}$ to constrain adaptation for subsequent tools.
For the next tool, we set $U_{<k+1}=U_{\leq k}$ and recompute the available projector using \cref{eq:available_tool_space}.
This procedure progressively reserves tool-selection-sensitive directions while allowing subsequent tools to adapt within the remaining shared tool-call directions.

%% file: Sections/4.Experiments.tex
\section{Experiments}
\label{sec:experiments}

\input{Tables/tab_perception_spatial}
\input{Tables/tab_ocr_depth}

\subsection{Experimental Setup}
\label{sec:experimental_setup}

\paragrapht{Agent and tool space.}
    We use DeepEyesV2~\cite{hong2026deepeyesv2}, built on Qwen2.5-VL-7B~\cite{bai2025qwen25vl}, as our primary post-trained multimodal agent and report additional results on Metis~\cite{yan2026metis} in \cref{sec:ablation} and Appendix.
    We sequentially extend the agent with three frozen vision specialists accessed through executable interfaces:
    SVTRv2~\cite{Du2025SVTRv2} for scene text recognition (OCR), GroundingDINO~\cite{liu2024groundingdino} for open-set object detection, and DepthAnything~\cite{yang2024depthanything} for monocular depth estimation.

\paragrapht{Training data.}
    Following \cref{sec:expandable_tools}, we generate multi-turn trajectories from annotated datasets for the corresponding tasks and retain only rollouts verified by rule-based checks.
    Each verified set is randomly partitioned into disjoint SFT and RL splits: ICDAR 2015~\cite{karatzas2015icdar} yields 1,463 SFT and 166 RL examples for OCR, Ref-Adv-s~\cite{dong2026refadv} yields 739 SFT and 87 RL examples for object detection, and UniQA-3D~\cite{zuo2025towards} yields 834 SFT and 77 RL examples for relative depth.
    Every retained trajectory contains intermediate reasoning, tool invocation, returned tool observations, and a verified final answer.

\paragrapht{Calibration set.}
    For the initial shared tool-call and protected tool-selection bases (\cref{sec:orthogonal_extension}), we compute the decoupled sensitivities in \cref{eq:subspace_sensitivity} over publicly released tool-use trajectories of the frozen initial agent.
    During sequential extension, selection-sensitive directions for each newly learned tool are estimated on its verified SFT trajectories.
    All calibration samples are disjoint from benchmarks.

\paragrapht{Benchmarks.}
    We evaluate four capability groups in the main text: (1) high-resolution visual perception on V*Bench~\cite{wu2024vstar} and HRBench~\cite{wang2025hrbench} to test retention of the native cropping tool, (2) text-rich visual understanding on SEED-Bench-2-Plus~\cite{li2024seedbench2plus}, TextVQA~\cite{singh2019textvqa}, and OCRBench~\cite{liu2024ocrbench}, (3) spatial reasoning and visual grounding on TreeBench~\cite{wang2025treebench}, MultihopSpatial~\cite{lee2026multihopspatial}, and MC-Bench~\cite{xu2025mcbench}, and (4) relative-depth perception on BLINK~\cite{fu2024blink} and multi-point HardBLINK~\cite{bigverdi2025perception}.
    Retention on eight additional mathematical reasoning, visual search, and factuality VQA benchmarks is reported in the Appendix.
    All evaluations follow each benchmark's standard protocol and metric.


\paragrapht{Baselines.}
    We compare \ours against training-free In-Context Prompting and three sequential LoRA adaptation methods.
    Prompting prepends tool schemas and verified demonstrations to the input prompt while keeping all model weights frozen.
    Naive SFT sequentially trains unconstrained per-tool LoRA adapters on the verified trajectories.
    O-LoRA~\cite{wang2023olora} penalizes subspace overlap between new and prior adapters, whereas KeepLoRA~\cite{luo2026keeplora} projects new updates into the residual subspace orthogonal to pre-trained weight principal components and the dominant feature directions of previous tasks.
    All adaptation baselines share identical training trajectories, LoRA ranks, and optimization budgets.

\begin{figure*}[t!]
    \centering
    \vspace{0pt}
    \includegraphics[width=\linewidth]{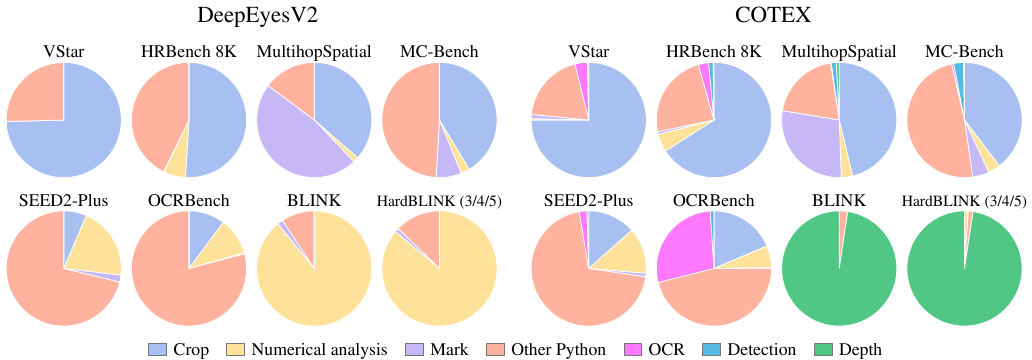}
    \vspace{-1.5em}
    \caption{
    Distribution of tool calls on each benchmark before and after training with \ours.
    }
    \label{fig:tool_distribution}\vspace{-1em}
\end{figure*}

\paragrapht{Implementation details.}
    We sequentially train tool-specific LoRA adapters for OCR, object detection, and depth estimation, each with rank $r=64$ and scaling factor $\alpha=128$.
    During SFT, each adapter is trained for one epoch with a constant learning rate of $1\times10^{-6}$, a global batch size of 8, and a maximum sequence length of 4,096 tokens.
    We then apply HDPO~\cite{yan2026metis} with group size $N=16$, learning rate $1\times10^{-6}$, reward weights $w_{\mathrm{acc}}=0.9$ and $w_{\mathrm{fmt}}=0.1$, and efficiency weight $w_{\mathrm{eff}}=0.15$.
    RL rollouts use temperature 1.0, top-$p$ 1.0, unrestricted top-$k$ ($k=-1$), up to 10 interaction turns, and a 300-second tool timeout.

\subsection{Main Results}
\label{sec:main_results}

\paragrapht{Visual perception and spatial reasoning.}
\cref{tab:perception} presents visual perception and spatial reasoning performance on six evaluation settings.
On benchmarks requiring the existing cropping tool (\ie, VStar, HRbench), \ours maintains or slightly improves performance relative to the original agent.
These results demonstrate that \ours effectively preserves the visual perception capabilities of the original agent.
Moreover, compared with DeepEyesV2, \ours consistently improves performance across all three spatial reasoning benchmarks, which require object grounding. 
\ours especially achieves gains of 5.0 and 12.2 points on MultihopSpatial and MC-Bench, respectively.
Interestingly, we empirically found that the substantial gain on MC-Bench is driven by the detection tool resolving challenging samples that the original agent failed to solve.

\paragrapht{Text-oriented understanding.}
The left part of \cref{tab:document} reports performance on text-oriented multimodal benchmarks.
Compared with DeepEyesV2, \ours improves OCRBench accuracy by 2.1 points, from 80.4 to 82.5, with increased OCR tool usage as shown in \cref{fig:tool_distribution}, while also improving TextVQA accuracy from 76.6 to 76.9 and maintaining comparable performance on SEED-Bench-2-Plus.
These results indicate that \ours retains previously acquired text-understanding capabilities while benefiting from the newly introduced OCR tool.

\paragrapht{Relative-depth perception.}
\cref{tab:document} presents relative-depth reasoning performance on BLINK and HardBLINK.
Compared with DeepEyesV2, \ours substantially improves BLINK accuracy from 41.1 to 87.1.
More notably, \ours achieves accuracies of 95.2, 85.5, and 88.7 on the three-, four-, and five-point HardBLINK benchmarks, respectively, outperforming DeepEyesV2 by 64.5, 64.5, and 69.3 points.
These substantial improvements demonstrate the effectiveness of incorporating specialized depth estimation into multimodal reasoning, particularly for challenging multi-point depth comparisons.

\subsection{Tool Call Distribution}
\cref{fig:tool_distribution} shows how \ours changes tool selection across benchmarks.
On OCRBench, the agent adopts the OCR tool while reducing the proportion of calls to other Python operations.
The shift is particularly pronounced on BLINK and HardBLINK, where depth estimation accounts for nearly all tool calls after training, replacing the previous reliance on numerical analysis.
Meanwhile, crop remains the most frequently invoked tool on VStar and HRBench 8K, and mark continues to account for a substantial share of calls on MultihopSpatial.
These patterns suggest that \ours incorporates new specialist tools selectively across benchmarks while continuing to use existing tools.

\begin{figure*}[h]
    \centering
    \begin{subfigure}[t]{0.32\linewidth}
        \centering
        \vspace{0pt}
        \includegraphics[width=\linewidth]{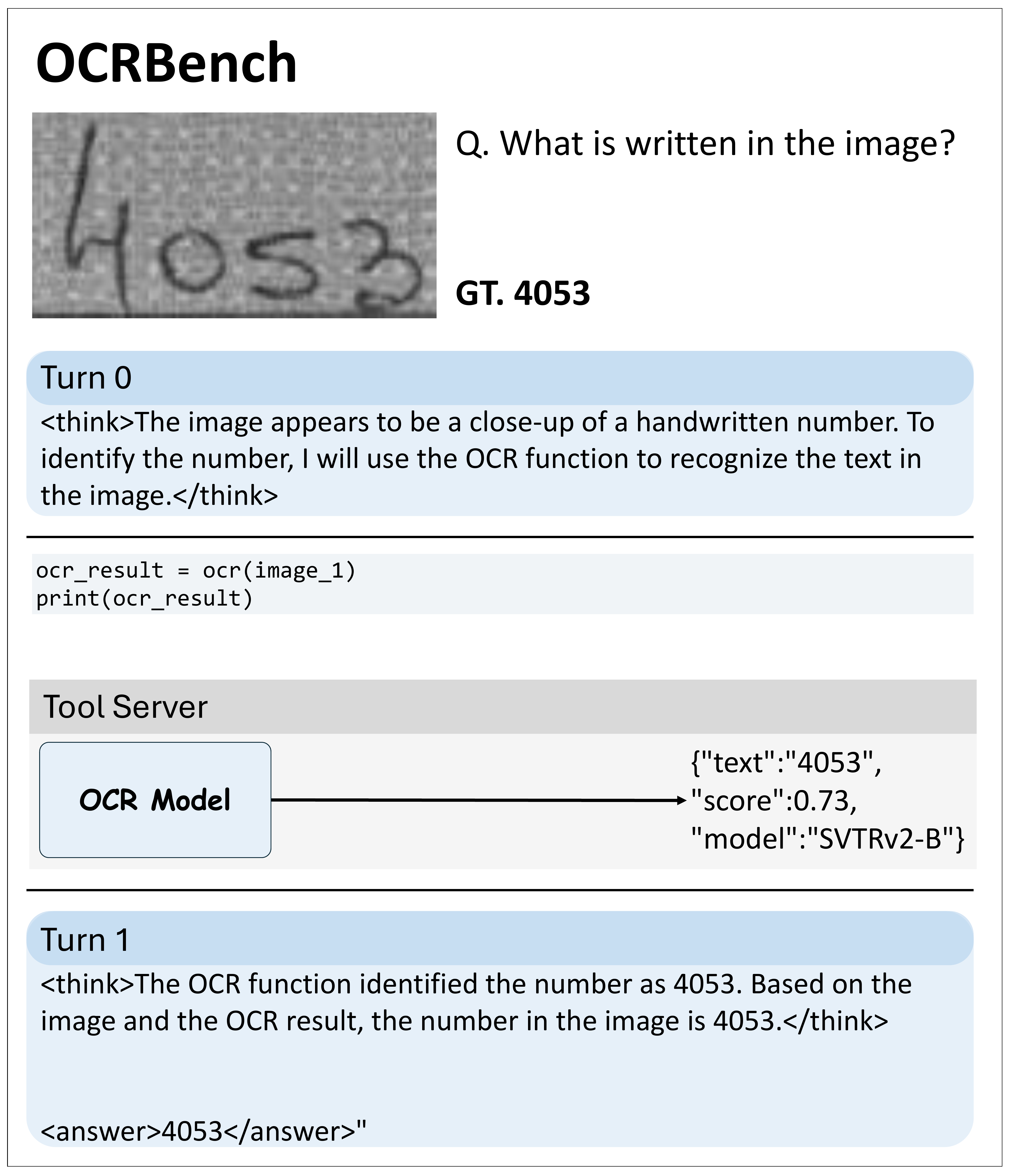}
        \caption{OCR}
        \label{fig:qual_ocr}
    \end{subfigure}
    \hfill
    \begin{subfigure}[t]{0.32\linewidth}
        \centering
        \vspace{0pt}
        \includegraphics[width=\linewidth]{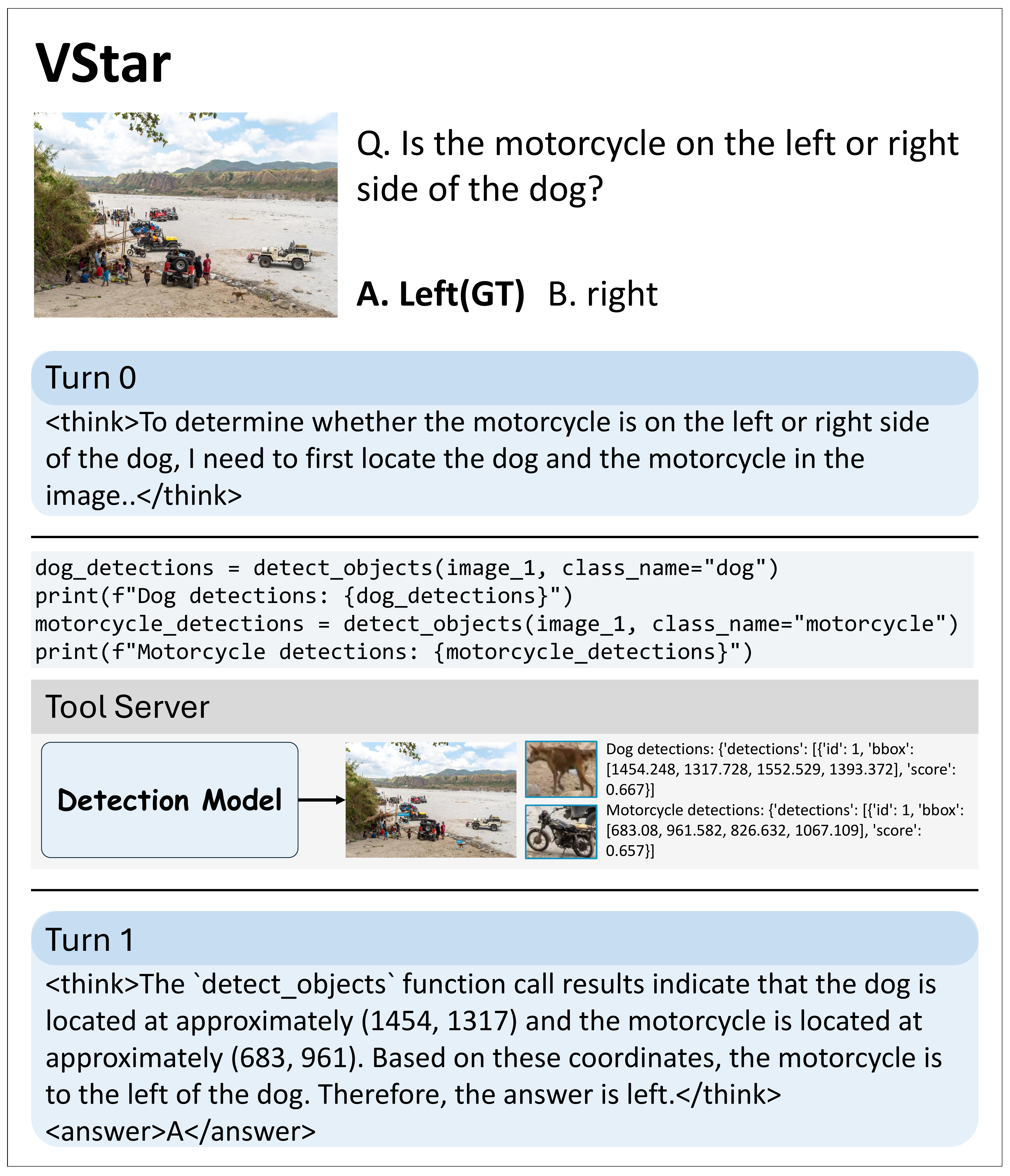}
        \caption{Object detection}
        \label{fig:qual_detection}
    \end{subfigure}
    \hfill
    \begin{subfigure}[t]{0.32\linewidth}
        \centering
        \vspace{0pt}
        \includegraphics[width=\linewidth]{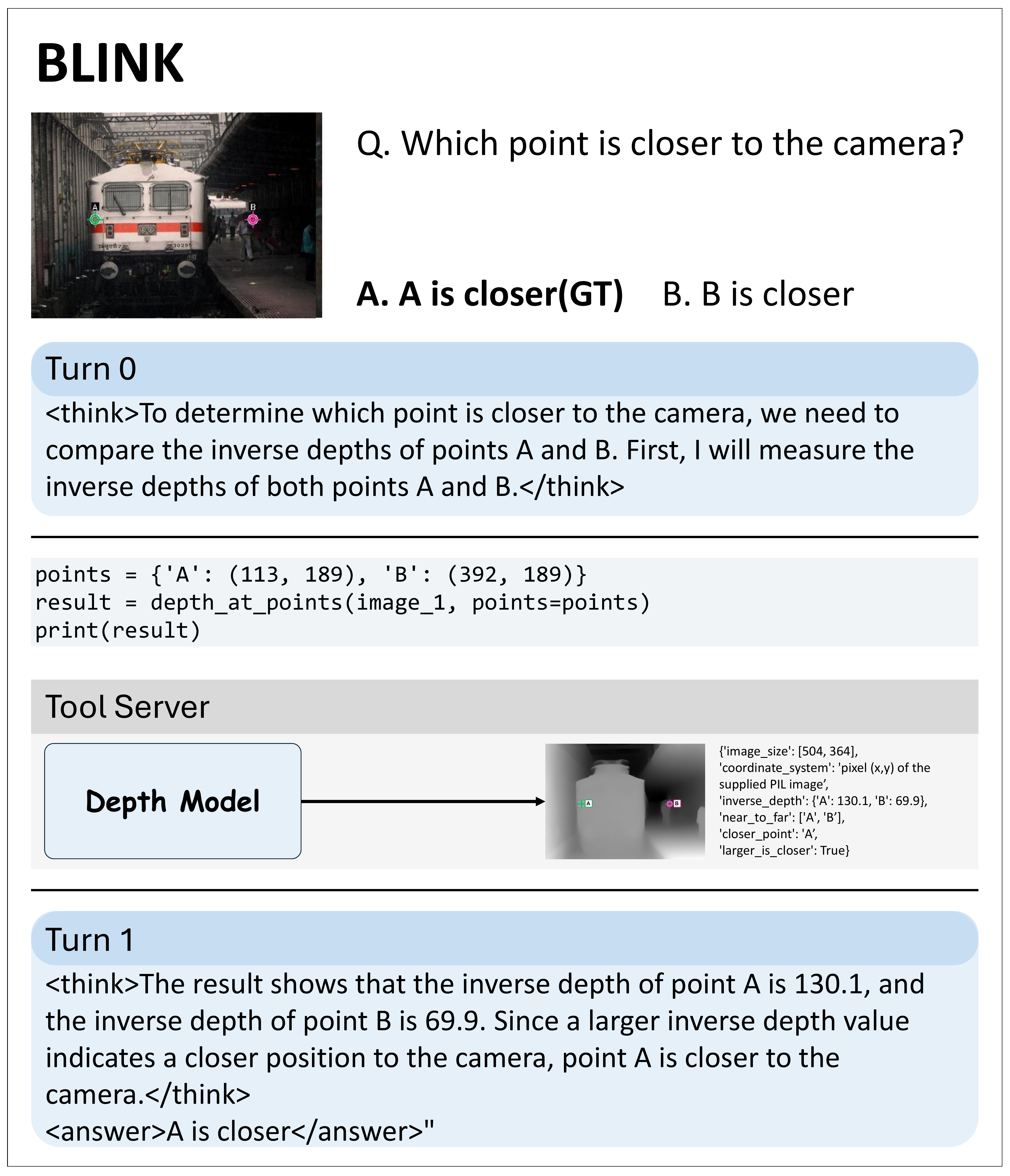}
        \caption{Depth estimation}
        \label{fig:qual_depth}
    \end{subfigure}
    \vspace{-.7em}
    \caption{
        Qualitative examples of newly introduced tool capabilities.
        Each example illustrates the agent's reasoning, tool invocation, returned observation, and final prediction.
        More examples are in Appendix.
    }
    \label{fig:tool_qualitative}\vspace{-.7em}
\end{figure*}

\subsection{Ablation Studies}
\label{sec:ablation}

\input{Tables/tab_ablation_method}

\paragrapht{Adaptation methods and sequential extension.}
\cref{tab:continual_extension} compares \ours with Naive SFT, O-LoRA~\cite{wang2023olora}, and KeepLoRA~\cite{luo2026keeplora} under sequential tool-space extension.
All methods use identical tool-use trajectories, LoRA configurations, and optimization budgets.
Across all extension stages, \ours achieves the highest V*Bench performance, retaining 80.6 after integrating all three tools, compared with 75.9, 78.0, and 74.3 for Naive SFT, O-LoRA, and KeepLoRA, respectively.
For newly acquired capabilities, \ours achieves the highest OCR accuracy of 46.4 after the first extension and the highest depth accuracy of 93.5 after the final extension, surpassing KeepLoRA by 6.5 points on depth estimation.
Although detection accuracy decreases after the depth extension, these results demonstrate the effectiveness of \ours in preserving general visual perception while acquiring specialized capabilities through sequential adaptation.

\input{Tables/tab_ablation_base_model_main}

\paragrapht{Different initial agents.}
To investigate the generalizability of \ours, we apply our framework to two post-trained multimodal agents, DeepEyesV2~\cite{hong2026deepeyesv2} and Metis~\cite{yan2026metis}.
As shown in \cref{tab:initial_agent_main}, \ours improves V*Bench accuracy by 1.6 points for DeepEyesV2 while maintaining comparable HRBench performance.
Similarly, applying \ours to Metis improves V*Bench accuracy from 90.6 to 92.5 and HRBench accuracy by 1.6 and 0.9 points at 4K and 8K resolutions, respectively.
These results demonstrate that \ours can extend different post-trained agents while preserving their previously acquired visual perception capabilities.

\input{Tables/tab_ablation_stage}

\paragrapht{SFT \& RL.}
We further analyze the contributions of SFT tool integration and subsequent RL in \cref{tab:rl_ablation}.
Following SFT, RL improves VStar accuracy up to 83.3, surpassing the initial agent's performance of 81.7.
On the task-specific examples reserved for RL training, optimization further improves OCR and detection accuracies by 1.8 and 11.5 points, respectively.
In particular, detection tool invocation increases from 32.2\% to 82.8\%, accompanied by an accuracy improvement from 35.6 to 47.1.
In contrast, depth accuracy decreases from 93.5 to 74.0 despite an unchanged tool invocation rate of 100.0\%.
These results show that RL improves tool utilization and general visual perception performance.
\vspace{-0.2em}

\subsection{Qualitative Results}
\label{sec:qualitative_results}

\cref{fig:tool_qualitative} presents representative tool-use trajectories for OCR, object detection, and depth estimation.
Each example shows how the agent invokes a specialized tool to acquire additional visual evidence and incorporates its observations into the final prediction.
These examples complement the quantitative results by demonstrating how newly introduced tools support task-specific multimodal reasoning.

%% file: Tables/tab_perception_spatial.tex
\begin{table*}[t]
    \centering
    \small
    \caption{
        Performance comparisons on visual perception benchmarks, including VStar~\cite{wu2024vstar} and HRBench~\cite{wang2025hrbench}, and spatial reasoning benchmarks, including TreeBench~\cite{wang2025treebench}, MultihopSpatial~\cite{lee2026multihopspatial}, and MC-Bench~\cite{xu2025mcbench}.
        The best and second-best scores in each column are shown in bold and underlined, respectively.
    }
    \label{tab:perception}\vspace{-5pt}
    \begin{tabular}{l ccc ccc}
        \toprule
        Method
        & VStar & HRbench4K & HRbench8K
        & TreeBench & MultihopSpatial & MC-Bench \\
        \cmidrule(lr){1-1} \cmidrule(lr){2-4} \cmidrule(lr){5-7}

        \rowcolor{gray!8} Qwen2.5-VL
        & 75.3 & 65.5 & 62.1
        & 37.0 & 2.9 & 14.7 \\

        DeepEyesV2
        & 81.7 & \textbf{76.6} & \underline{71.3}
        & \underline{43.2} & \underline{12.4} & 38.4 \\

        \, + Prompting
        & \underline{82.2} & 73.5 & 69.3
        & 41.7 & 10.2 & 27.1 \\

        \, + Finetuning
        & 75.9 & 74.9 & 70.4
        & 40.5 & 10.8 & \underline{45.3} \\

        \, + \ours
        & \textbf{83.3} & \underline{76.3} & \textbf{71.4}
        & \textbf{43.5} & \textbf{17.4} & \textbf{50.6} \\

        \bottomrule
    \end{tabular}
\end{table*}

%% file: Tables/tab_ocr_depth.tex
\begin{table*}[t]
    \centering
    \small
    \caption{
            Performance comparisons on text-rich VL benchmarks, including SEED2Plus~\cite{li2024seedbench2plus}, TextVQA~\cite{singh2019textvqa}, OCRBench~\cite{liu2024ocrbench} and depth perception benchmarks, including BLINK~\cite{fu2024blink}, HardBLINK~\cite{bigverdi2025perception}. `HB(\#points)' denotes the number of comparison points in HardBLINK.
    }\vspace{-5pt}
    \label{tab:document}
    \begin{tabular}{l ccc cccc}
        \toprule
        Method
        & SEED2Plus & TextVQA
        & OCRBench & BLINK & HB(3) & HB(4) & HB(5) \\
        \cmidrule(lr){1-1} \cmidrule(lr){2-4} \cmidrule(lr){5-8}

        \rowcolor{gray!8} Qwen2.5-VL
        & 68.8 & 72.0 & 80.6 & 47.6 & 37.1 & 29.0 & 15.3 \\

        DeepEyesV2
        & \bf 69.8 & 76.6 & 80.4 & 41.1 & 30.7 & 21.0 & 19.4 \\

        \, + Prompting
        & 69.3 & \underline{76.7} & 80.1
        & \underline{62.1} & 38.7 & 27.4 & \underline{26.6} \\

        \, + Finetuning
        & \underline{69.4} & \underline{76.7} & \underline{80.9}
        & 58.1 & \underline{43.5} & \underline{32.3} & 25.0 \\

        \, + \ours
        & \underline{69.4} & \textbf{76.9} & \textbf{82.5}
        & \textbf{87.1} & \textbf{95.2} & \textbf{85.5} & \textbf{88.7} \\

        \bottomrule
    \end{tabular}
\end{table*}

%% file: Tables/tab_ablation_method.tex
\begin{table}[t]
    \centering
    \small
    \setlength{\tabcolsep}{4pt}
    \caption{
        Comparison of adaptation methods under sequential tool-space extension.
        Each stage cumulatively introduces OCR (+O), detection (+G), and depth (+D).
        All values are accuracies (\%).
    }\vspace{-5pt}
    \label{tab:continual_extension}
    \begin{tabular}{llcccc}
        \toprule
        Method & Stage & VStar & OCR & Detection & Depth \\
        \midrule

        DeepEyesV2
        & Initial & 81.7 & -- & -- & -- \\

        \midrule

        \multirow{3}{*}{Finetuning}
        & +O & 79.1 & 31.9 & --   & --   \\
        & +G & 75.9 & 41.6 & 50.6 & --   \\
        & +D & 75.9 & 44.0 & 47.1 & 64.9 \\

        \midrule

        \multirow{3}{*}{O-LoRA}
        & +O & 79.1 & 31.9 & --   & --   \\
        & +G & 75.4 & 39.8 & 50.6 & --   \\
        & +D & 78.0 & 42.2 & 51.7 & 64.9 \\

        \midrule

        \multirow{3}{*}{KeepLoRA}
        & +O & 79.6 & 40.4 & --   & --   \\
        & +G & 74.9 & 36.7 & 54.0 & --   \\
        & +D & 74.3 & 43.4 & \textbf{56.3} & 87.0 \\

        \midrule

        \multirow{3}{*}{\ours}
        & +O & 81.7 & \textbf{46.4} & --   & --   \\
        & +G & \textbf{83.8} & 45.8 & 50.6 & --   \\
        & +D & 80.6 & 42.2 & 35.6 & \textbf{93.5} \\
        \bottomrule
    \end{tabular}
    \vspace{-1em}
\end{table}

%% file: Tables/tab_ablation_base_model_main.tex
\begin{table}[t]
    \centering
    \small
    \caption{
        Effect of before/after \ours with different initial agents. Performance on other benchmarks is in Appendix.
    }
    \label{tab:initial_agent_main}\vspace{-5pt}
    \begin{tabular}{llccc}
        \toprule
        Initial Agent
        & Method
        & VStar
        & HR4K
        & HR8K \\
        \midrule

        \multirow{2}{*}{DeepEyesV2}
        & Initial
        & 81.7 & \textbf{76.6} & 71.3 \\

        & \ours
        & \textbf{83.3} & 76.3 & \textbf{71.4} \\

        \midrule

        \multirow{2}{*}{Metis}
        & Initial
        & 90.6 & 79.0 & 78.6 \\

        & \ours
        & \textbf{92.5} & \textbf{80.6} & \textbf{79.5} \\

        \bottomrule
    \end{tabular}\vspace{-1em}
\end{table}

%% file: Tables/tab_ablation_stage.tex
\begin{table}[t]
    \centering
    \small
    \caption{
        Effectiveness on SFT and RL from the official DeepEyesV2 weights. Call denotes the tool invocation rate (\%).
    }
    \vspace{-5pt}
    \label{tab:rl_ablation}
    \begin{tabular}{lc cc cc cc}
        \toprule
        \multirow{2}{*}{Stage}
        & \multirow{2}{*}{VStar}
        & \multicolumn{2}{c}{OCR}
        & \multicolumn{2}{c}{Detection}
        & \multicolumn{2}{c}{Depth} \\
        \cmidrule(lr){3-4}
        \cmidrule(lr){5-6}
        \cmidrule(lr){7-8}
        &
        & Acc. & Call
        & Acc. & Call
        & Acc. & Call \\
        \midrule

        Initial
        & 81.7
        & -- & --
        & -- & --
        & -- & -- \\

        + SFT
        & 80.6
        & 42.2 & 28.9
        & 35.6 & 32.2
        & 93.5 & 100.0 \\

        + RL
        & 83.3
        & 44.0 & 38.0
        & 47.1 & 82.8
        & 74.0 & 100.0 \\

        \bottomrule
    \end{tabular}
    \vspace{-1.5em}
\end{table}

%% file: Sections/5.Conclusion.tex
\vspace{-0.3em}

\section{Conclusion}
\label{sec:conclusion}
\vspace{-0.1em}
We introduce \ours, a continual tool space extension learning framework for scalable multimodal agents.
\ours recovers two sets of directions to derive the initiation of a tool call, and to determine which tool is chosen.
Each new tool is learned as a low-rank adapter inside the former and orthogonal to the latter.
Therefore, the shared invocation ability is reused while existing tool choices are left intact.
Its training data come from existing annotated datasets for each tasks, with no teacher model involved.
In our experiments, the agents gained the new capabilities with almost no loss on the tools they already used.

%% file: Supple/A.Training.tex
\section{Algorithm of \ours}
\label{app:algorithm}

Algorithm~\ref{alg:orthogonal_sft} summarizes the overall training procedure of \ours, from shared and protected subspace construction to sequential orthogonal low-rank adaptation and policy optimization.
The shared tool-call basis is estimated from calibration trajectories of the initial agent, whereas each newly learned tool contributes additional protected tool-selection directions.
During adaptation, previously learned adapters remain active but frozen, and only the current tool-specific low-rank parameters $\{A, B\}$ are updated.

\input{Tables/supple/tab_algorithm}

\section{More Experimental Details}
\label{app:experimental_details}
\label{app:training_details}

\input{Tables/supple/tab_hyperparms}

\subsection{Training Configuration}
We conduct experiments on two multimodal agents:  DeepEyesV2~\cite{hong2026deepeyesv2} and Metis~\cite{yan2026metis},
based on Qwen2.5-VL~\cite{bai2025qwen25vl} and Qwen3-VL~\cite{bai2025qwen3vl}, respectively.
We employ new orthogonal parameters upon the projection layers of the LLM decoder, keeping all pretrained weights frozen (\eg, vision encoder, multimodal connector, LLM,
and external specialist models).
The detailed training recipes for SFT and RL are provided in \cref{tab:supp_hparams}, with the corresponding training scripts (\texttt{SFT/train.py} and \texttt{RL/train.py}) included under both \texttt{code/DeepEyesV2/} and \texttt{code/Metis/} in the supplementary code.

\subsection{Verification and Data Construction}
\input{Tables/supple/tab_data}
We construct verified tool-use trajectories from existing task-specific annotations, using ICDAR 2015~\cite{karatzas2015icdar} for OCR, Ref-Adv-s~\cite{dong2026refadv} for object detection, and UniQA-3D~\cite{zuo2025towards} for relative depth.
For OCR, we retain trajectories whose transcriptions exactly match the ICDAR 2015 annotations after Unicode NFKC normalization and whitespace trimming, with case-sensitive matching.
For object detection, we retain trajectories when the target category is unambiguous and exactly one Grounding DINO~\cite{liu2024groundingdino} detection has an IoU of at least 0.5 with the target bounding box.
For depth estimation, we require valid coordinates and finite inverse-depth values at both points, retaining trajectories whose predicted depth ordering agrees with the UniQA-3D annotations.
Larger inverse-depth values indicate closer points, and we balance the retained trajectories between closer and farther questions.

\cref{tab:supp_data_stats} summarizes the number of generated candidates and selected trajectories, including their disjoint SFT and RL partitions.
Across the three tools, 3,366 trajectories are selected from 4,080 candidates (82.5\% selection rate), yielding 3,036 SFT and 330 RL examples.
The RL partition is used for validation after SFT and subsequent policy optimization, rather than as an independent held-out test set.

\subsection{Tool Interfaces}
\label{sec:suppl_tool_interfaces}

Each specialist model remains frozen and is accessed through an executable interface, with tool observations appended to the agent's interaction history.
The OCR interface returns recognized text and associated model outputs; the detector accepts an image and a text query and returns object detections; and the depth interface returns relative-depth estimates for queried pixel locations.
We include the tool-specific prompt files for OCR (\texttt{ocr\_template.json}), detection (\texttt{detection\_template.json}), and depth estimation (\texttt{depth\_template.json}) in each model's \texttt{prompts/} directory.

\paragrapht{Prompts.}
We use separate prompts for agent inference and verified tool-use trajectory generation.
The agent interaction prompt follows each Agents' format~\cite{hong2026deepeyesv2, yan2026metis} and is shown in Section~\ref{sec:prompts}. \Cref{prompt:agent_deepeyesv2,prompt:agent_metis}.
The shared trajectory-generation template is shown in \Cref{prompt:trajectory_deepeyesv2,prompt:trajectory_metis}. 
We attached tool-specific prompt files for OCR, detection, and depth estimation in code.

\begin{figure*}[h]
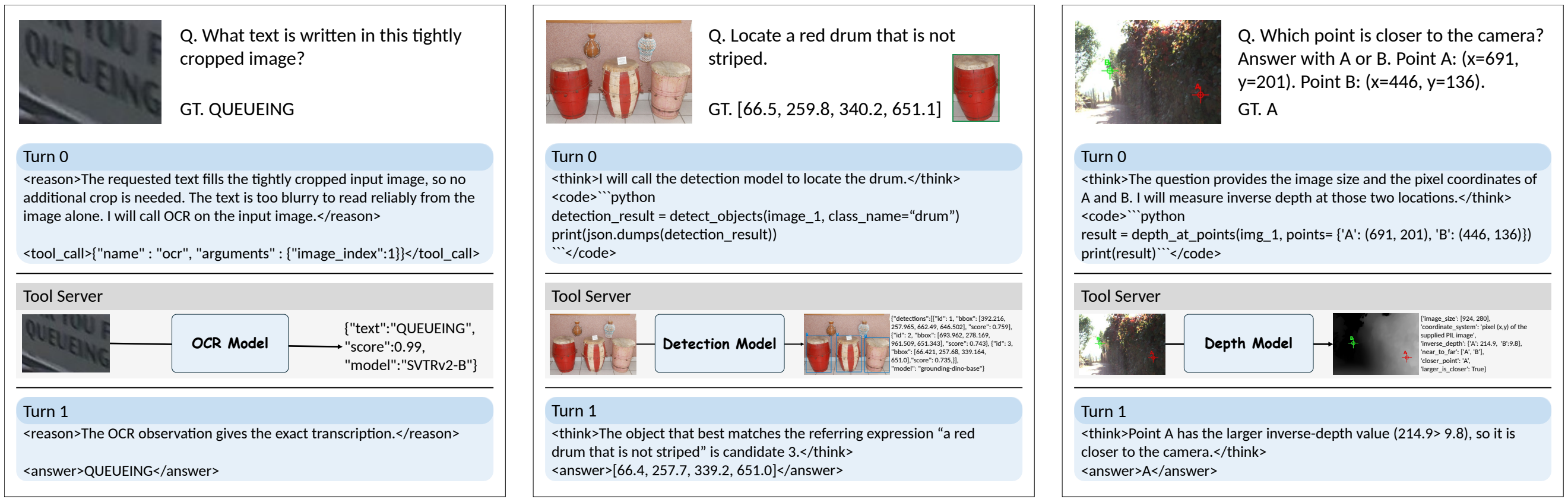

    \centering
    \begin{subfigure}[t]{0.32\linewidth}
        \centering
        \includegraphics[width=\linewidth]{Figure/Supple/example_ocr.pdf}
        \caption{OCR}
        \label{fig:supp_example_ocr}
    \end{subfigure}
    \hfill
    \begin{subfigure}[t]{0.32\linewidth}
        \centering
        \includegraphics[width=\linewidth]{Figure/Supple/example_detection.pdf}
        \caption{Object detection}
        \label{fig:supp_example_detection}
    \end{subfigure}
    \hfill
    \begin{subfigure}[t]{0.32\linewidth}
        \centering
        \includegraphics[width=\linewidth]{Figure/Supple/example_depth.pdf}
        \caption{Depth estimation}
        \label{fig:supp_example_depth}
    \end{subfigure}
    \vspace{-.7em}
    \caption{Examples of verified tool-use trajectories generated for supervised training. Each example illustrates the agent's reasoning, tool invocation, returned observation, and final prediction.}
    \label{fig:supp_training_examples}
\end{figure*}

\paragrapht{Training examples.}\label{app:training_examples}
\cref{fig:supp_training_examples} presents representative verified training trajectories for OCR, detection, and depth estimation.
These examples illustrate how existing task annotations are converted into agent-generated tool-use supervision through task-specific verification.

%% file: Tables/supple/tab_algorithm.tex
\renewcommand{\algorithmicrequire}{\textbf{Input:}}
\renewcommand{\algorithmicensure}{\textbf{Output:}}
\begin{algorithm*}[h]
\caption{The overall training procedure of \ours.}
\label{alg:orthogonal_sft}
\begin{algorithmic}[1]




\Require Base parameters $\theta^\star$, agent parameters $\theta$, a set of calibration trajectories $\{i\}$ of $\mathcal{T}_{\mathrm{old}}$, 
\Statex \hspace{\algorithmicindent} verified new tool trajectories $\{\mathcal{D}_k\}_{k=1}^{K}$ with SFT and RL splits, dimensions $d_{\mathrm{call}}$ and low-rank $r$
\Ensure Adapters $\{(B_k, A_k)\}_{k=1}^{K}$, protected basis $U_{<K+1}$
\Statex \textit{Stage 1: Subspace identification} \hfill (Sec.3.3)
\State $\Delta W \gets W_{\theta} - W_{\theta^\star}$
\State $U_\Delta \gets$ left singular vectors of $\Delta W$
\State $U_{\mathrm{call}} \gets U_\Delta\, Q_{[d_{\mathrm{call}}]}\big(C(U_\Delta, \mathcal{L}_{\mathrm{call}})\big)$ over $\{i\}$ \Comment{shared tool-call basis}
\State $U_{\mathrm{old}} \gets U_{\mathrm{call}}\, Q_{[r]}\big(C(U_{\mathrm{call}}, \mathcal{L}_{\mathrm{tool}})\big)$ over $\{i\}$, \; $U_{<1} \gets U_{\mathrm{old}}$ \Comment{protected basis}
\Statex \textit{Stage 2: Continual tool-space learning} \hfill (Secs.3.4, 3.5)
\For{$k = 1$ to $K$}
    \State $P_k \gets U_{\mathrm{call}} U_{\mathrm{call}}^{\top} - U_{<k} U_{<k}^{\top}$ \Comment{available adaptation space}
    \State $R_{<k} \gets$ orthonormal basis of $\operatorname{range}(P_k)$
    \State Attach adapter $\Delta W_k = s_k R_{<k} B_k A_k$ \Comment{$\theta$ and $\{\Delta W_j\}_{j<k}$ frozen}
    \State Update $(B_k, A_k)$ by SFT on the SFT split of $\mathcal{D}_k$
    \State Update $(B_k, A_k)$ by RL with $\mathcal{L}_{\mathrm{RL}}$ on the RL split of $\mathcal{D}_k$
    \State $U_k \gets R_{<k}\, Q_{[r]}\big(C(R_{<k}, \mathcal{L}_{\mathrm{tool},k})\big)$ over $\mathcal{D}_k$ \Comment{selection directions of $g_k$}
    \State $U_{\le k} \gets [\,U_{<k},\ U_k\,]$, \; $U_{<k+1} \gets U_{\le k}$
\EndFor
\State \Return $\{(B_k, A_k)\}_{k=1}^{K}$, $U_{<K+1}$

\end{algorithmic}
\end{algorithm*}

%% file: Tables/supple/tab_hyperparms.tex
\begin{table}[t]
    \centering
    \small
    \setlength{\tabcolsep}{4pt}
    \caption{Training configurations for COTEX.}
    \label{tab:supp_hparams}
    \begin{tabular*}{0.5\columnwidth}{@{\extracolsep{\fill}}lc@{}}
        \toprule
        \multicolumn{2}{@{}l}{\textbf{Common settings}} \\
        \midrule
        Learning rate & $1\times10^{-6}$ \\
        Optimizer & AdamW \\
        Learning-rate schedule & Constant \\
        Random seed & 23 \\
        Numerical precision & BF16 \\
        Gradient norm clipping & 1.0 \\
        Weight decay & 0 \\
        Training epochs & 1 \\
        Microbatch size per GPU & 1 \\
        GPU model / memory & NVIDIA RTX A6000 / 48\,GB \\
        \# of training GPUs & 8 \\

        \midrule
        \multicolumn{2}{@{}l}{\textbf{SFT}} \\
        \midrule
        LoRA rank & 64 \\
        LoRA scaling $\alpha$ & 128 \\
        Global batch size & 8 \\
        Maximum sequence length & 4,096 \\
        Gradient accumulation & 1 \\

        \midrule
        \multicolumn{2}{@{}l}{\textbf{RL}} \\
        \midrule
        Rollouts per prompt & 16 \\
        Accuracy reward weight $w_{\mathrm{acc}}$ & 0.9 \\
        Format reward weight $w_{\mathrm{fmt}}$ & 0.1 \\
        Efficiency weight $w_{\mathrm{eff}}$ & 0.15 \\
        Temperature & 1.0 \\
        Top-$p$ & 1.0 \\
        Top-$k$ & $-1$ \\
        Agent turns & 10 \\
        Tool timeout & 300\,s \\
        PPO clipping $\epsilon$ & 0.2 \\
        KL regularization & 0 \\
        Rollout backend & vLLM \\
        \bottomrule
    \end{tabular*}
\end{table}

%% file: Tables/supple/tab_data.tex
\begin{table*}[t]
    \centering
    \small
    \setlength{\tabcolsep}{5pt}
    \caption{Candidate and selected trajectories for tool integration. Selection rate is the percentage of candidate trajectories retained for SFT or RL. OCR counts include both scene-crop and direct variants.}
    \label{tab:supp_data_stats}
    \begin{tabular}{ll cccc c}
        \toprule
        Tool & Source dataset & Candidates & Selected & SFT & RL & Selection rate (\%) \\
        \midrule
        OCR       & ICDAR 2015 & 1,938 & 1,629 & 1,463 & 166 & 84.06 \\
        Detection & Ref-Adv-s  & 1,142 &   826 &   739 &  87 & 72.33 \\
        Depth     & UniQA-3D   & 1,000 &   911 &   834 &  77 & 91.10 \\
        \midrule
        Total     & --         & 4,080 & 3,366 & 3,036 & 330 & 82.50 \\
        \bottomrule
    \end{tabular}
\end{table*}

%% file: Supple/B.Exp.tex
\section{Additional Experiments}
\label{app:additional_experiments}

\subsection{Mathematical Reasoning Tasks}
\label{app:reasoning_search}

We further investigate whether \ours preserves the mathematical reasoning capabilities of the initial agent.
\cref{tab:reasoning_search} presents performance comparisons on five mathematical and multimodal reasoning benchmarks, including MathVista~\cite{lu2023mathvista}, MathVerse~\cite{zhang2024mathverse}, WeMath~\cite{qiao2024wemath}, DynaMath~\cite{zou2024dynamath}, and LogicVista~\cite{xiao2024logicvista}.
All models are evaluated with Python as an available tool.
We employ Qwen3.5-122B-A10B-FP8~\cite{qwen2026qwen35} as an LLM-based judge to assess the correctness of the predicted answers.
Compared with DeepEyesV2, \ours slightly improves MathVista accuracy from 71.1 to 71.2 and maintains the same accuracy of 46.9 on LogicVista, with performance drops of at most 1.6 points across the remaining evaluation settings.
Moreover, \ours outperforms both O-LoRA~\cite{wang2023olora} and KeepLoRA~\cite{luo2026keeplora} on five out of seven evaluation settings and matches KeepLoRA on the worst-case metric of DynaMath.
These results demonstrate the effectiveness of \ours in mitigating interference with previously acquired mathematical reasoning capabilities while incorporating new specialized visual capabilities.

\input{Tables/tab_reasoning}

\subsection{Different Initial Agents}
\label{app:base_model}

\input{Tables/supple/tab_metis_perception_spatial}
\input{Tables/supple/tab_metis_ocr_depth}

To further investigate the generalizability of \ours, we extend Metis~\cite{yan2026metis} with the same three specialist tools used in our main experiments with DeepEyesV2~\cite{hong2026deepeyesv2}.
\cref{tab:supp_metis_visual,tab:supp_metis_text_depth} present the performance comparisons across visual perception, spatial reasoning, text-oriented understanding, and relative-depth perception benchmarks.

\paragrapht{Visual perception and spatial reasoning.}
As shown in \cref{tab:supp_metis_visual}, \ours improves VStar accuracy from 90.6 to 92.5 and HRBench accuracy by 1.6 and 0.9 points at 4K and 8K resolutions, respectively.
Consistent with our observations on DeepEyesV2, these results demonstrate that \ours effectively preserves previously acquired visual perception capabilities across different initial agents.
On spatial reasoning benchmarks, \ours improves MultihopSpatial accuracy from 38.2 to 38.5 while maintaining comparable performance on TreeBench and MC-Bench.
Notably, \ours consistently outperforms naive finetuning across all six evaluation settings, achieving gains of 4.7 and 21.8 points on MultihopSpatial and MC-Bench, respectively.
These results demonstrate the effectiveness of orthogonal adaptation in mitigating interference with previously acquired visual perception and grounding capabilities.

\paragrapht{Text-oriented understanding.}
\cref{tab:supp_metis_text_depth} presents performance comparisons on text-rich vision-language benchmarks.
Compared with Metis, \ours improves TextVQA and OCRBench accuracies by 1.4 and 0.6 points, respectively, while maintaining the same performance on SEED-Bench-2-Plus.
In particular, \ours achieves gains of 6.0 and 5.5 points over naive finetuning on TextVQA and OCRBench, respectively.
These results indicate that \ours effectively integrates the OCR tool while preserving previously acquired text-understanding capabilities.

\paragrapht{Relative-depth perception.}
Compared with Metis, \ours substantially improves BLINK accuracy from 55.7 to 98.4.
More notably, \ours achieves accuracies of 98.4, 96.0, and 97.6 on the three-, four-, and five-point HardBLINK benchmarks, respectively, outperforming Metis by 52.6, 58.0, and 64.5 points.
Furthermore, \ours consistently matches or surpasses naive finetuning across all relative-depth evaluation settings.
Together with our observations on DeepEyesV2, these results demonstrate the effectiveness of \ours in incorporating specialized depth estimation capabilities across different post-trained multimodal agents.

\section{More Qualitative Results}
\label{app:qualitative_results}

We provide additional qualitative comparisons on DeepEyesV2 and Metis to illustrate how newly introduced specialist tools complement the capabilities of different initial agents.
\cref{fig:supp_corrected_all} presents examples in which the initial agents produce incorrect predictions, whereas \ours correctly answers the same questions by acquiring additional visual evidence through newly introduced tools.
Each example shows the reasoning trajectories, tool invocations, returned observations, and final predictions before and after tool-space extension.

\paragrapht{OCR.}
In \cref{fig:supp_corrected_ocr}, DeepEyesV2 incorrectly recognizes the stylized text as ``GATNER'' even after inspecting the image through Python execution.
In contrast, \ours invokes the OCR tool and correctly identifies the text as ``GAFFNER'' using the returned transcription.
This example illustrates how specialized text recognition complements the visual perception capabilities of the initial agent.

\paragrapht{Object detection.}
In \cref{fig:supp_corrected_detection}, DeepEyesV2 incorrectly identifies the table as the target object and estimates its bounding box based on visual inspection.
After tool-space extension, \ours invokes the detection tool to locate the flowerpot and subsequently converts the returned bounding box into normalized coordinates through Python execution.
This example illustrates how \ours incorporates newly introduced detection capabilities into the existing multi-turn reasoning process by combining specialist and computational tools for visual grounding.

\paragrapht{Depth estimation.}
In \cref{fig:supp_corrected_depth}, Metis incorrectly identifies point B as the closest point among four marked points based on their vertical image coordinates.
In contrast, \ours invokes the depth estimation tool to obtain the inverse-depth values of all four points and correctly identifies point A as the closest.
Notably, although the depth training trajectories involve two-point comparisons, \ours successfully utilizes the depth tool for multi-point comparisons on HardBLINK.
This example illustrates how the newly acquired tool-use capability generalizes beyond the interaction scenarios encountered during training.

These qualitative results demonstrate that \ours integrates specialized perceptual capabilities into the existing multi-turn reasoning processes of different post-trained agents, allowing them to resolve challenging examples they previously failed to solve.

\begin{figure*}[p]
    \centering
    \begin{subfigure}[t]{\linewidth}
        \centering
        \includegraphics[width=\linewidth]{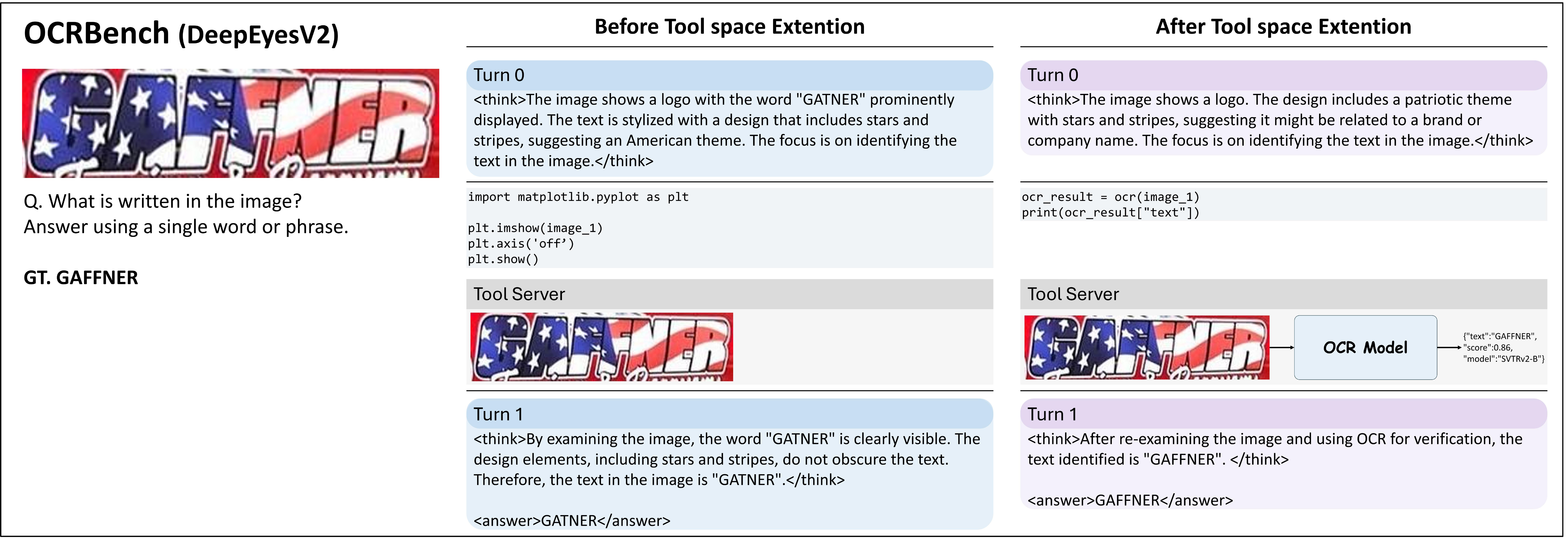}
        \caption{OCR}
        \label{fig:supp_corrected_ocr}
    \end{subfigure}
\end{figure*}

\begin{figure*}[p]
    \ContinuedFloat
    \centering
    \begin{subfigure}[t]{\linewidth}
        \centering
        \includegraphics[width=\linewidth,height=0.85\textheight,keepaspectratio]{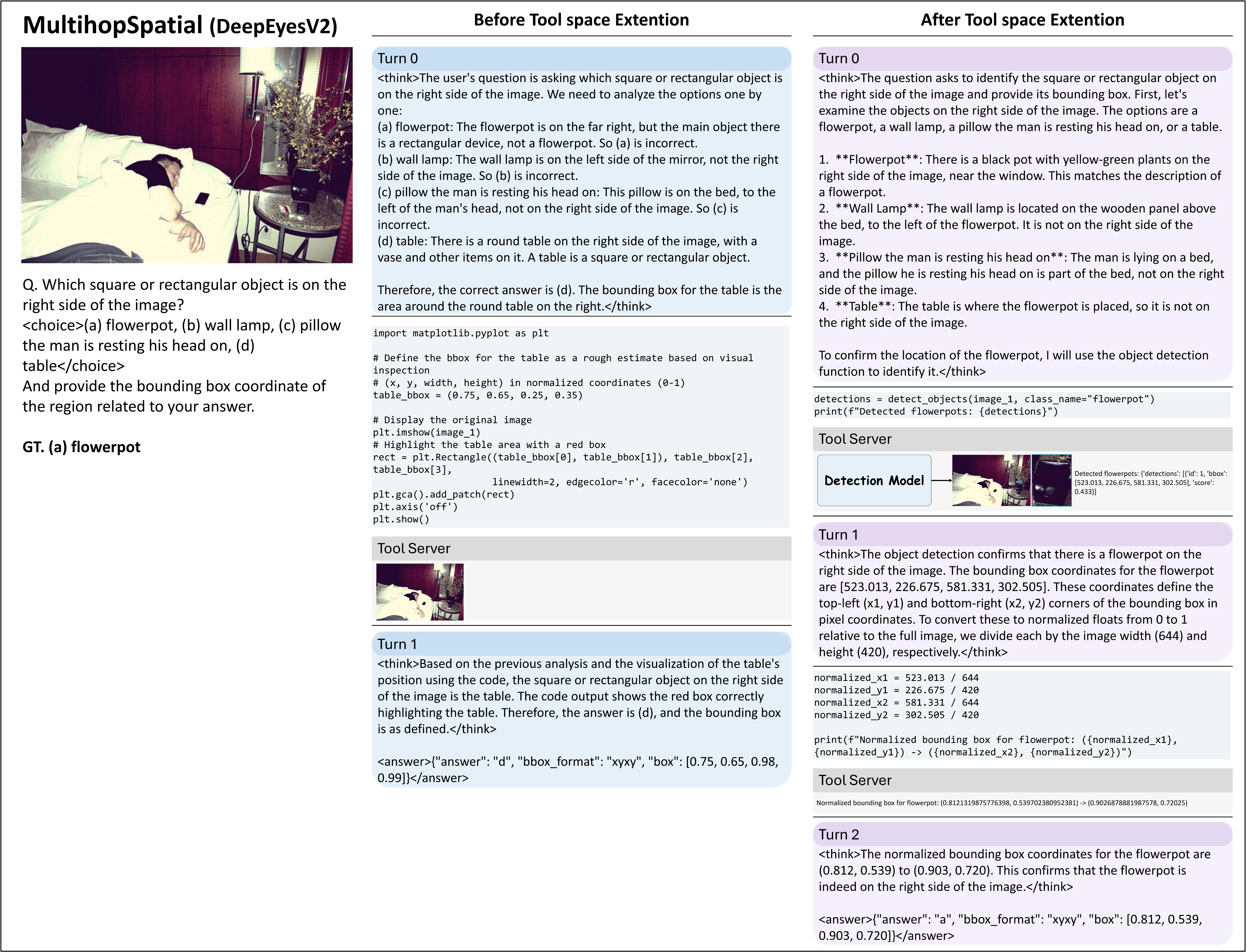}
        \caption{Object detection}
        \label{fig:supp_corrected_detection}
    \end{subfigure}
\end{figure*}

\begin{figure*}[p]
    \ContinuedFloat
    \centering
    \begin{subfigure}[t]{\linewidth}
        \centering
        \includegraphics[width=\linewidth,height=0.85\textheight,keepaspectratio]{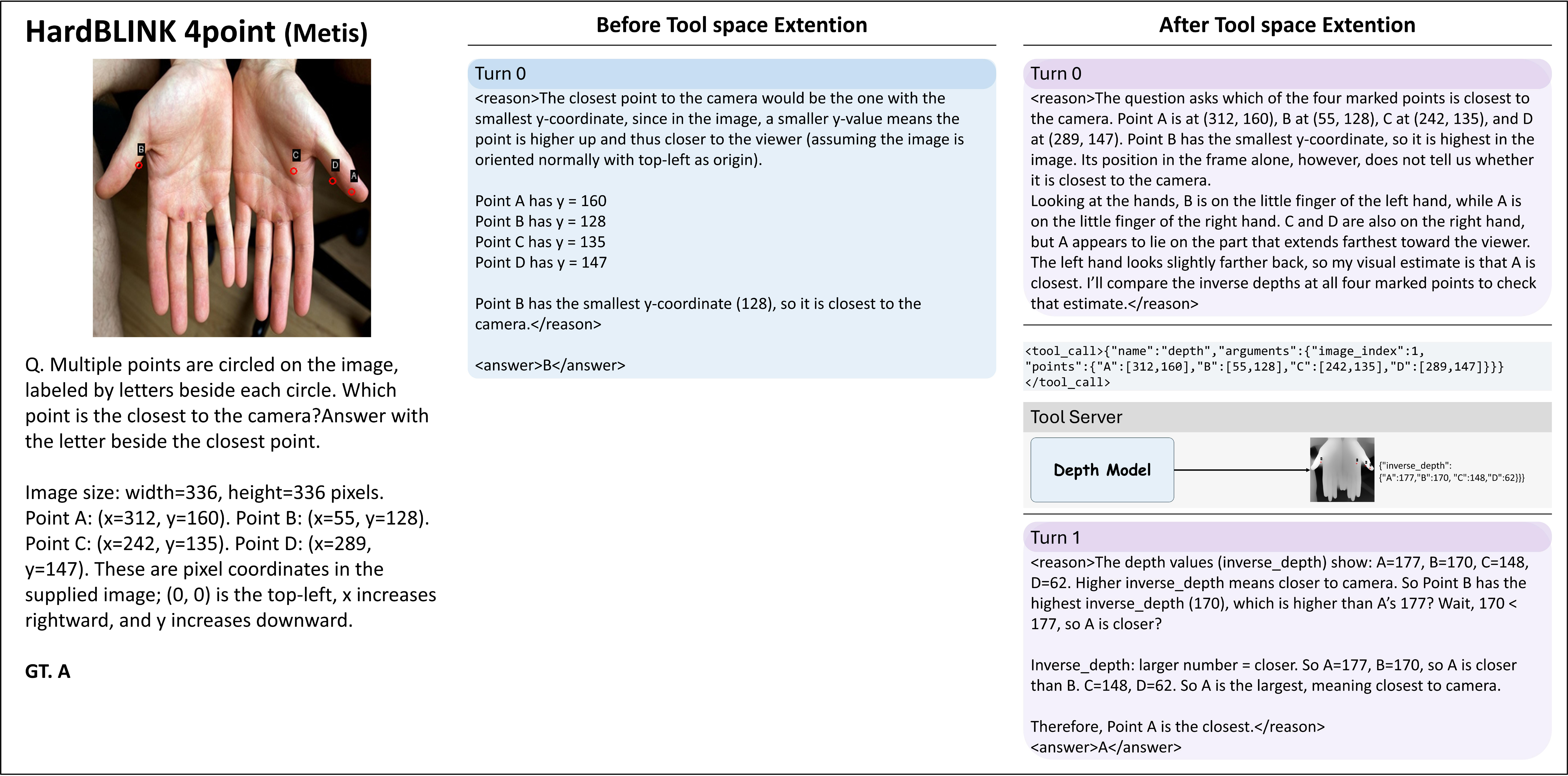}
        \caption{Depth estimation}
        \label{fig:supp_corrected_depth}
    \end{subfigure}
    \caption{
        Qualitative comparisons before and after tool-space extension.
        Each example illustrates how \ours corrects an incorrect prediction of the initial agent by incorporating observations from newly introduced specialist tools.
    }
    \label{fig:supp_corrected_all}
\end{figure*}



%% file: Tables/tab_reasoning.tex
\begin{table*}[h]
    \centering
    \small
    \caption{
        Performance comparison on mathematical reasoning benchmarks.
        All models are evaluated with Python as an available tool.
        The best and second-best scores for each metric are shown in bold and underlined, respectively.
    }
    \label{tab:reasoning_search}
    \begin{tabular}{lccccc}
        \toprule
        \multirow{2}{*}{Model}
        & \multirow{2}{*}{MathVista MINI}
        & \multirow{2}{*}{MathVerse MINI}
        & WeMath & DynaMath
        & \multirow{2}{*}{LogicVista} \\
        & & & Strict / Loose & Avg. / Worst & \\
        \midrule

        Qwen2.5-VL
        & 67.6 & 47.3
        & 32.8 / 52.6
        & 54.5 / 21.8
        & \underline{46.4} \\

        DeepEyesV2
        & \underline{71.1} & \textbf{52.0}
        & \textbf{37.4} / \textbf{57.6}
        & \textbf{58.0} / \textbf{24.8}
        & \textbf{46.9} \\

        O-LoRA
        & \underline{71.1} & 50.5
        & 33.9 / 55.1
        & 57.3 / 22.4
        & 44.0 \\

        KeepLoRA
        & 69.7 & 50.1
        & 34.0 / 55.9
        & \underline{57.9} / \underline{23.6}
        & 46.2 \\

        \ours{}
        & \textbf{71.2} & \underline{50.7}
        & \underline{36.1} / \underline{56.0}
        & 57.4 / \underline{23.6}
        & \textbf{46.9} \\

        \bottomrule
    \end{tabular}
    \par\smallskip
\end{table*}

%% file: Tables/supple/tab_metis_perception_spatial.tex
\begin{table*}[t]
    \centering
    \small
    \caption{
        Performance comparisons on visual perception benchmarks, including VStar~\cite{wu2024vstar} and HRBench~\cite{wang2025hrbench}, and spatial reasoning benchmarks, including TreeBench~\cite{wang2025treebench}, MultihopSpatial~\cite{lee2026multihopspatial}, and MC-Bench~\cite{xu2025mcbench}.
    }
    \label{tab:supp_metis_visual}\vspace{-5pt}
    \begin{tabular}{l ccc ccc}
        \toprule
        Method
        & VStar & HRbench4K & HRbench8K
        & TreeBench & MultihopSpatial & MC-Bench \\
        \cmidrule(lr){1-1} \cmidrule(lr){2-4} \cmidrule(lr){5-7}

        \rowcolor{gray!8} Qwen3-VL
        & 85.3 & 78.1 & 75.3
        & 40.0 & 32.2 & 26.6 \\

        Metis
        & \underline{90.6} & 79.0 & \underline{78.6}
        & \textbf{46.2} & \underline{38.2} & \textbf{63.4} \\

        \, + Prompting
        & 88.5 & 76.8 & 74.1
        & \underline{45.9} & 36.7 & 61.5 \\

        \, + Finetuning
        & 85.3 & \underline{80.0} & 77.3
        & 43.2 & 33.8 & 39.8 \\

        \, + \ours
        & \textbf{92.5} & \textbf{80.6} & \textbf{79.5}
        & \underline{45.9} & \textbf{38.5} & \underline{61.6} \\

        \bottomrule
    \end{tabular}
\end{table*}

%% file: Tables/supple/tab_metis_ocr_depth.tex
\begin{table*}[t]
    \centering
    \small
    \caption{
        Performance comparisons on text-rich VL benchmarks, including SEED2Plus~\cite{li2024seedbench2plus}, TextVQA~\cite{singh2019textvqa}, OCRBench~\cite{liu2024ocrbench}, and depth perception benchmarks, including BLINK~\cite{fu2024blink} and HardBLINK~\cite{bigverdi2025perception}.
        `HB(\#points)' denotes the number of comparison points in HardBLINK.
    }
    \label{tab:supp_metis_text_depth}\vspace{-5pt}
    \begin{tabular}{l ccc cccc}
        \toprule
        Method
        & SEED2Plus & TextVQA & OCRBench
        & BLINK & HB(3) & HB(4) & HB(5) \\
        \cmidrule(lr){1-1} \cmidrule(lr){2-4} \cmidrule(lr){5-8}

        \rowcolor{gray!8} Qwen3-VL
        & 71.0 & 71.5 & 80.3
        & 17.7 & 27.4 & 22.6 & 19.4 \\

        Metis
        & \textbf{71.9} & 76.4 & \underline{84.0}
        & \underline{55.7} & 45.8 & 38.0 & 33.1 \\

        \, + Prompting
        & \underline{71.6} & \textbf{78.0} & 83.1
        & 51.6 & 36.3 & 26.6 & 24.2 \\

        \, + Finetuning
        & \textbf{71.9} & 71.8 & 79.1
        & \textbf{98.4} & \underline{96.8} & \underline{95.2} & \underline{96.8} \\

        \, + \ours
        & \textbf{71.9} & \underline{77.8} & \textbf{84.6}
        & \textbf{98.4} & \textbf{98.4} & \textbf{96.0} & \textbf{97.6} \\

        \bottomrule
    \end{tabular}
\end{table*}

%% file: Supple/C.Limitation.tex
\section{Limitation}\label{sec:limitation}
Since \ours derives the shared tool-call subspace from the displacement between an agent and its base MLLM, it is applicable when both sets of parameters are accessible and the base model is known.
Agents distributed without this information, or served only through an API, fall outside the scope of our method.
Identifying the tool-call subspace from the agent’s weights alone, for example through activation statistics on tool-call tokens, is a promising direction to relax this requirement.

%% file: Supple/prompt.tex
\clearpage
\section{Prompt Templates}
\label{sec:prompts}

\begin{rtppromptbox}[label={prompt:agent_deepeyesv2}]{Agent Model (DeepEyesV2)}
You are a multimodal problem-solving agent. Answer the user's question accurately using the image and any tools that provide useful evidence.

# Tools
You have access to a Python environment. The current input image is available as image_1.

Use Python to inspect image dimensions, crop a region, or call one of the following functions:
- ocr(image): Read text from an image or crop. The recognized text is available in the result's "text" field.
- detect_objects(image, class_name): Find objects of the requested class. The result contains candidate detections and bounding boxes.
- depth_at_points(image, points): Measure inverse depth at labeled pixel coordinates. The result's "inverse_depth" field maps each label to its measured value.

# Decision Guidelines
1. For exact transcription, identify the target text. If it occupies part of a larger image, crop that region before calling ocr.
2. For referring-expression localization, call detect_objects with the target class. Use the description to select the matching candidate and return its bounding box in [x1, y1, x2, y2] pixel coordinates.
3. For marked-point depth comparisons, use the supplied coordinates when available. Otherwise, inspect the image dimensions and locate the markers. Measure both points in one call. Larger inverse depth means closer to the camera.
4. For other questions, answer directly when the visual evidence is sufficient. Use Python when a specific observation would help.

# Output Format
Begin each turn with <think>...</think>. Then either request one Python execution or give the final answer.

Example tool request for a tightly cropped text image:
<think>I need the exact lettering on this label.</think>
<code>```python
result = ocr(image_1)
print(result["text"])
```</code>

After receiving the observation, use it to answer or request another operation if needed. Give the final answer in the requested format:
<think>The observation resolves the question.</think>
<answer>...</answer>
\end{rtppromptbox}

\begin{rtppromptbox}[label={prompt:agent_metis}]{Agent Model (Metis)}
You are a multimodal problem-solving agent. Answer the user's question accurately using visual evidence and the available tools.

# Tools
- python: Call with a "code" string. The current input image is available as image_1.

Use print() to return values and plt.show() to return an image. Variables and imports persist within the session.
- ocr: Call with a one-based integer "image_index". The observation contains "text" and "score".

Original input images occupy the first indices. Images returned by tools are appended in creation order.
- object_detection: Call with "backend" set to "grounding_dino", "image_index", a one-category "class_filter", "confidence_threshold" set to 0.35, and "return_boxes" set to true.

The observation contains "image_size" and a list of detections with "id", "bbox", and "score".
- depth: Call with "image_index" and "points". Map each marker label to its [x, y] pixel coordinates.

The observation contains "inverse_depth" values for the requested points.

# Decision Guidelines
1. For exact transcription, call ocr on the target text. If the input already tightly isolates the text, use its image index directly. Otherwise, crop the target with python, return the crop, and call ocr with the returned image's index.
2. For referring-expression localization, call object_detection with the target class. Compare the candidates with the description and return the selected candidate's bbox.
3. For marked-point depth comparisons, use the supplied coordinates when available. Otherwise, inspect the image dimensions and locate the markers. Pass both points in one depth call. Larger inverse depth means closer to the camera.
4. For other questions, choose a tool when it can resolve a specific uncertainty. Answer directly when the visual evidence is sufficient.

# Output Format
Begin each turn with <reason>...</reason>. Then emit exactly one JSON tool call or a final answer. Use the returned observation as evidence on the next turn.

Example tool request for a tightly cropped text image:
<reason>I need the exact lettering on this label.</reason>
<tool_call>{"name":"ocr","arguments":{"image_index":1}}</tool_call>

Example final-answer format:
<reason>The observation resolves the question.</reason>
<answer>...</answer>
\end{rtppromptbox}

\begin{rtppromptbox}[label={prompt:trajectory_deepeyesv2}]{Tool-use Trajectory Template (DeepEyesV2)}
[SYSTEM]
You are a multimodal [TASK] agent. Use the available Python functions to obtain the evidence needed to answer the user's question.

# Tools
[INSERT THE PYTHON FUNCTION SIGNATURES AND TASK GUIDELINES AVAILABLE AT THE CURRENT STAGE.]

The current question image is available as image_1. Python variables persist across execution turns, so a crop stored in one turn can be passed directly to a tool function in a later turn.

# Output Format
Begin each assistant turn with <think>...</think>. Then emit one Python code block inside <code>...</code> or give the final answer inside <answer>...</answer>.

[USER]
<image>
[QUESTION CONSTRUCTED FROM THE SOURCE DATASET]

[ASSISTANT — OPTIONAL PREPROCESSING]
<think>[BRIEF PURPOSE OF THE IMAGE PREPROCESSING STEP]</think>
<code>```python
[EXECUTABLE PYTHON CODE THAT INSPECTS THE INPUT OR STORES THE TARGET REGION IN A NAMED VARIABLE]
```</code>

[OBSERVATION — USER ROLE]
Code execution result:
stdout: ```[ACTUAL STANDARD OUTPUT, IF ANY] ```
stderr: ```[ACTUAL STANDARD ERROR, IF ANY] ```
Images: [RETURNED IMAGE, IF ANY]

[ASSISTANT — TARGET TOOL]
<think>[BRIEF PURPOSE OF APPLYING THE TARGET FUNCTION TO THE SELECTED IMAGE OR REGION]</think>
<code>```python
result = [TARGET_FUNCTION]([IMAGE_VARIABLE], [ADDITIONAL_ARGUMENTS_IF_REQUIRED])
print([RELEVANT_RESULT_FIELD])
```</code>

[OBSERVATION — USER ROLE]
Code execution result:
stdout: ```[ACTUAL PRINTED TOOL RESULT]```
stderr: ```[ACTUAL STANDARD ERROR, IF ANY] ```
Images: [RETURNED IMAGE, IF ANY]

[ASSISTANT — FINAL ANSWER]
<think>[BRIEF SUMMARY OF THE OBSERVATION THAT SUPPORTS THE ANSWER]</think>
<answer>[GROUND-TRUTH ANSWER IN THE FORMAT REQUESTED BY THE QUESTION]</answer>
\end{rtppromptbox}

\begin{rtppromptbox}[label={prompt:trajectory_metis}]{Tool-use Trajectory Template (Metis)}
[SYSTEM]
You are a multimodal [TASK] agent. Use the available tools to obtain the evidence needed to answer the user's question.

# Tools
[INSERT THE TOOL DESCRIPTIONS, JSON ARGUMENT SCHEMAS, AND TASK GUIDELINES AVAILABLE AT THE CURRENT STAGE.]

Image indices are one-based. Original input images occupy the first indices, and images returned by tools are appended in creation order. When a tool request needs a crop returned by Python, pass the image_index assigned to that crop.

# Output Format
Begin each assistant turn with <reason>...</reason>. Then emit one JSON tool call inside <tool_call>...</tool_call> or give the final answer inside <answer>...</answer>.

[USER]
<image>
[QUESTION CONSTRUCTED FROM THE SOURCE DATASET]

[ASSISTANT — OPTIONAL PREPROCESSING]
<reason>[BRIEF PURPOSE OF THE IMAGE PREPROCESSING STEP]</reason>
<tool_call>{"name":"python","arguments":{"code":"[EXECUTABLE PYTHON CODE THAT INSPECTS THE INPUT OR RETURNS THE TARGET REGION]"}}</tool_call>

[OBSERVATION — USER ROLE]
[ACTUAL PYTHON EXECUTION OUTPUT]
[RETURNED IMAGE, IF ANY]

[ASSISTANT — TARGET TOOL]
<reason>[BRIEF PURPOSE OF CALLING THE TARGET TOOL ON THE SELECTED IMAGE OR REGION]</reason>
<tool_call>{"name":"[TARGET_TOOL]","arguments":[ARGUMENT OBJECT CONFORMING TO THE TARGET TOOL SCHEMA]}</tool_call>

[OBSERVATION — USER ROLE]
[ACTUAL JSON OBSERVATION RETURNED BY THE TARGET TOOL]

[ASSISTANT — FINAL ANSWER]
<reason>[BRIEF SUMMARY OF THE OBSERVATION THAT SUPPORTS THE ANSWER]</reason>
<answer>[GROUND-TRUTH ANSWER IN THE FORMAT REQUESTED BY THE QUESTION]</answer>
\end{rtppromptbox}


%% file: main.bbl
\begin{thebibliography}{43}
\providecommand{\natexlab}[1]{#1}
\providecommand{\url}[1]{\texttt{#1}}
\expandafter\ifx\csname urlstyle\endcsname\relax
  \providecommand{\doi}[1]{doi: #1}\else
  \providecommand{\doi}{doi: \begingroup \urlstyle{rm}\Url}\fi

\bibitem[Bai et~al.(2025)Bai, Chen, Liu, Wang, Ge, Song, Dang, Wang, Wang, Tang, et~al.]{bai2025qwen25vl}
Shuai Bai, Keqin Chen, Xuejing Liu, Jialin Wang, Wenbin Ge, Sibo Song, Kai Dang, Peng Wang, Shijie Wang, Jun Tang, et~al.
\newblock {Qwen2.5-VL} technical report.
\newblock \emph{arXiv preprint arXiv:2502.13923}, 2025.

\bibitem[Bigverdi et~al.(2025)Bigverdi, Luo, Hsieh, Shen, Chen, Shapiro, and Krishna]{bigverdi2025perception}
Mahtab Bigverdi, Zelun Luo, Cheng-Yu Hsieh, Ethan Shen, Dongping Chen, Linda~G. Shapiro, and Ranjay Krishna.
\newblock Perception tokens enhance visual reasoning in multimodal language models.
\newblock In \emph{CVPR}, pages 3836--3845, 2025.

\bibitem[Dettmers et~al.(2023)Dettmers, Pagnoni, Holtzman, and Zettlemoyer]{dettmers2023qlora}
Tim Dettmers, Artidoro Pagnoni, Ari Holtzman, and Luke Zettlemoyer.
\newblock {QLoRA}: Efficient finetuning of quantized llms.
\newblock In \emph{NeurIPS}, 2023.

\bibitem[Dong et~al.(2026)Dong, Yang, Ju, Zhao, Zhang, Wang, Zeng, Lu, and Fu]{dong2026refadv}
Qihua Dong, Kuo Yang, Lin Ju, Handong Zhao, Yitian Zhang, Yizhou Wang, Huimin Zeng, Jianglin Lu, and Yun Fu.
\newblock {Ref-Adv}: Exploring {MLLM} visual reasoning in referring expression tasks.
\newblock In \emph{ICLR}, 2026.

\bibitem[Du et~al.(2025)Du, Chen, Xie, Jia, and Jiang]{Du2025SVTRv2}
Yongkun Du, Zhineng Chen, Hongtao Xie, Caiyan Jia, and Yu-Gang Jiang.
\newblock {SVTRv2}: {CTC} beats encoder-decoder models in scene text recognition.
\newblock In \emph{ICCV}, pages 20147--20156, 2025.

\bibitem[Fu et~al.(2024)Fu, Hu, Li, Feng, Wang, Lin, Roth, Smith, Ma, and Krishna]{fu2024blink}
Xingyu Fu, Yushi Hu, Bangzheng Li, Yu Feng, Haoyu Wang, Xudong Lin, Dan Roth, Noah~A. Smith, Wei-Chiu Ma, and Ranjay Krishna.
\newblock {BLINK}: Multimodal large language models can see but not perceive.
\newblock In \emph{ECCV}, 2024.

\bibitem[Gupta and Kembhavi(2023)]{gupta2023visprog}
Tanmay Gupta and Aniruddha Kembhavi.
\newblock Visual programming: Compositional visual reasoning without training.
\newblock In \emph{CVPR}, pages 14953--14962, 2023.

\bibitem[He et~al.(2026)He, Cheng, Wang, Yang, Wang, and Gao]{he2026loda}
Lingfeng He, De Cheng, Huaijie Wang, Xi Yang, Nannan Wang, and Xinbo Gao.
\newblock Task-driven subspace decomposition for knowledge sharing and isolation in lora-based continual learning.
\newblock In \emph{Proceedings of the 43rd International Conference on Machine Learning}, pages 41172--41194. PMLR, 2026.

\bibitem[Hong et~al.(2026)Hong, Zhao, Zhu, Lu, Xu, and Yu]{hong2026deepeyesv2}
Jack Hong, Chenxiao Zhao, Chenglin Zhu, Weiheng Lu, Guohai Xu, and Xing Yu.
\newblock {DeepEyesV2}: Toward agentic multimodal model.
\newblock In \emph{ICLR}, 2026.

\bibitem[Hu et~al.(2022)Hu, Shen, Wallis, Allen-Zhu, Li, Wang, Wang, and Chen]{hu2022lora}
Edward~J. Hu, Yelong Shen, Phillip Wallis, Zeyuan Allen-Zhu, Yuanzhi Li, Shean Wang, Lu Wang, and Weizhu Chen.
\newblock {LoRA}: Low-rank adaptation of large language models.
\newblock In \emph{ICLR}, 2022.

\bibitem[Huang et~al.(2024)Huang, Parthasarathi, Rezagholizadeh, and Chandar]{huang2024practical}
Jerry Huang, Prasanna Parthasarathi, Mehdi Rezagholizadeh, and Sarath Chandar.
\newblock Towards practical tool usage for continually learning {LLMs}.
\newblock \emph{arXiv preprint arXiv:2404.09339}, 2024.

\bibitem[Karatzas et~al.(2015)Karatzas, Gomez-Bigorda, Nicolaou, Ghosh, Bagdanov, Iwamura, Matas, Neumann, Chandrasekhar, Lu, Shafait, Uchida, and Valveny]{karatzas2015icdar}
Dimosthenis Karatzas, Lluis Gomez-Bigorda, Anguelos Nicolaou, Suman Ghosh, Andrew Bagdanov, Masakazu Iwamura, Jiri Matas, Lukas Neumann, Vijay~Ramaseshan Chandrasekhar, Shijian Lu, Faisal Shafait, Seiichi Uchida, and Ernest Valveny.
\newblock {ICDAR} 2015 competition on robust reading.
\newblock In \emph{International Conference on Document Analysis and Recognition}, pages 1156--1160, 2015.

\bibitem[Lee et~al.(2026)Lee, Jang, Cho, Lee, Lee, and Hwang]{lee2026multihopspatial}
Youngwan Lee, Soojin Jang, Yoorhim Cho, Seunghwan Lee, Yong-Ju Lee, and Sung~Ju Hwang.
\newblock {MultihopSpatial}: Multi-hop compositional spatial reasoning benchmark for vision-language model.
\newblock In \emph{ECCV}, pages 580--598, 2026.

\bibitem[Li et~al.(2024)Li, Ge, Chen, Ge, Zhang, and Shan]{li2024seedbench2plus}
Bohao Li, Yuying Ge, Yi Chen, Yixiao Ge, Ruimao Zhang, and Ying Shan.
\newblock {SEED-Bench-2-Plus}: Benchmarking multimodal large language models with text-rich visual comprehension.
\newblock \emph{arXiv preprint arXiv:2404.16790}, 2024.

\bibitem[Liu et~al.(2024{\natexlab{a}})Liu, Cheng, Liu, Zhang, Li, Ren, Zou, Yang, Su, Zhu, Zhang, Gao, and Li]{liu2024llavaplus}
Shilong Liu, Hao Cheng, Haotian Liu, Hao Zhang, Feng Li, Tianhe Ren, Xueyan Zou, Jianwei Yang, Hang Su, Jun Zhu, Lei Zhang, Jianfeng Gao, and Chunyuan Li.
\newblock {LLaVA-Plus}: Learning to use tools for creating multimodal agents.
\newblock In \emph{ECCV}, pages 126--142, 2024{\natexlab{a}}.

\bibitem[Liu et~al.(2024{\natexlab{b}})Liu, Zeng, Ren, Li, Zhang, Yang, Jiang, Li, Yang, Su, Zhu, and Zhang]{liu2024groundingdino}
Shilong Liu, Zhaoyang Zeng, Tianhe Ren, Feng Li, Hao Zhang, Jie Yang, Qing Jiang, Chunyuan Li, Jianwei Yang, Hang Su, Jun Zhu, and Lei Zhang.
\newblock {Grounding DINO}: Marrying {DINO} with grounded pre-training for open-set object detection.
\newblock In \emph{ECCV}, pages 38--55, 2024{\natexlab{b}}.

\bibitem[Liu et~al.(2024{\natexlab{c}})Liu, Li, Huang, Yang, Yu, Li, Yin, Liu, Jin, and Bai]{liu2024ocrbench}
Yuliang Liu, Zhang Li, Mingxin Huang, Biao Yang, Wenwen Yu, Chunyuan Li, Xu-Cheng Yin, Cheng-Lin Liu, Lianwen Jin, and Xiang Bai.
\newblock {OCRBench}: On the hidden mystery of ocr in large multimodal models.
\newblock \emph{Science China Information Sciences}, 67\penalty0 (12):\penalty0 220102, 2024{\natexlab{c}}.

\bibitem[Liu et~al.(2026)Liu, Cao, Shi, and Liang]{liu2025colt}
Yuyang Liu, Meng Cao, Xinyuan Shi, and Xiaodan Liang.
\newblock {COLT}: Enhancing video large language models with continual tool usage.
\newblock \emph{Transactions on Machine Learning Research}, 2026.

\bibitem[Luo et~al.(2026)Luo, Zhou, Zhang, Wan, Zhang, and Wei]{luo2026keeplora}
Mao-Lin Luo, Zi-Hao Zhou, Yi-Lin Zhang, Yuanyu Wan, Min-Ling Zhang, and Tong Wei.
\newblock Keeplora: Continual learning with residual gradient adaptation.
\newblock In \emph{ICLR}, 2026.

\bibitem[Marczak et~al.(2025)Marczak, Magistri, Cygert, Twardowski, Bagdanov, and van~de Weijer]{marczak2025notask}
Daniel Marczak, Simone Magistri, Sebastian Cygert, Bart{\l}omiej Twardowski, Andrew~D. Bagdanov, and Joost van~de Weijer.
\newblock No task left behind: Isotropic model merging with common and task-specific subspaces.
\newblock In \emph{Proceedings of the 42nd International Conference on Machine Learning}, pages 43177--43199. PMLR, 2025.

\bibitem[Shao et~al.(2024)Shao, Wang, Zhu, Xu, Song, Bi, Zhang, Zhang, Li, Wu, and Guo]{shao2024deepseekmath}
Zhihong Shao, Peiyi Wang, Qihao Zhu, Runxin Xu, Junxiao Song, Xiao Bi, Haowei Zhang, Mingchuan Zhang, Y.~K. Li, Y. Wu, and Daya Guo.
\newblock {DeepSeekMath}: Pushing the limits of mathematical reasoning in open language models.
\newblock \emph{arXiv preprint arXiv:2402.03300}, 2024.

\bibitem[Singh et~al.(2019)Singh, Natarajan, Shah, Jiang, Chen, Batra, Parikh, and Rohrbach]{singh2019textvqa}
Amanpreet Singh, Vivek Natarajan, Meet Shah, Yu Jiang, Xinlei Chen, Dhruv Batra, Devi Parikh, and Marcus Rohrbach.
\newblock Towards vqa models that can read.
\newblock In \emph{CVPR}, pages 8317--8326, 2019.

\bibitem[Sur{\'\i}s et~al.(2023)Sur{\'\i}s, Menon, and Vondrick]{suris2023vipergpt}
D{\'\i}dac Sur{\'\i}s, Sachit Menon, and Carl Vondrick.
\newblock {ViperGPT}: Visual inference via python execution for reasoning.
\newblock In \emph{ICCV}, pages 11888--11898, 2023.

\bibitem[Wang et~al.(2026{\natexlab{a}})Wang, Li, Huang, Wang, Wang, Zhang, Zheng, Bai, Kang, Feng, Wang, and Zhang]{wang2025treebench}
Haochen Wang, Xiangtai Li, Zilong Huang, Anran Wang, Jiacong Wang, Tao Zhang, Jiani Zheng, Sule Bai, Zijian Kang, Jiashi Feng, Zhuochen Wang, and Zhaoxiang Zhang.
\newblock Traceable evidence enhanced visual grounded reasoning: Evaluation and method.
\newblock In \emph{ICLR}, 2026{\natexlab{a}}.

\bibitem[Wang et~al.(2026{\natexlab{b}})Wang, Shi, Cheng, Zhang, He, Tang, Zhang, Yu, Chen, Xu, Zhang, Wan, Wang, and Ying]{wang2026beacon}
Qixun Wang, Yang Shi, Letian Cheng, Zhuoran Zhang, Yan He, Yuqi Tang, Qi Zhang, Xinlei Yu, Ruizhe Chen, Tianrun Xu, Yuanxing Zhang, Pengfei Wan, Haotian Wang, and Xianghua Ying.
\newblock Beacon: Knowing when and how to perform agentic visual reasoning.
\newblock \emph{arXiv preprint arXiv:2607.28595}, 2026{\natexlab{b}}.

\bibitem[Wang et~al.(2025)Wang, Ding, Zeng, Zhou, Shen, Luo, Yu, and Tao]{wang2025hrbench}
Wenbin Wang, Liang Ding, Minyan Zeng, Xiabin Zhou, Li Shen, Yong Luo, Wei Yu, and Dacheng Tao.
\newblock Divide, conquer and combine: A training-free framework for high-resolution image perception in multimodal large language models.
\newblock In \emph{AAAI}, pages 7907--7915, 2025.

\bibitem[Wang et~al.(2023)Wang, Chen, Ge, Xia, Bao, Zheng, Zhang, Gui, and Huang]{wang2023olora}
Xiao Wang, Tianze Chen, Qiming Ge, Han Xia, Rong Bao, Rui Zheng, Qi Zhang, Tao Gui, and Xuanjing Huang.
\newblock Orthogonal subspace learning for language model continual learning.
\newblock In \emph{Findings of EMNLP}, pages 10658--10671, 2023.

\bibitem[Wu and Xie(2024)]{wu2024vstar}
Penghao Wu and Saining Xie.
\newblock {V*}: Guided visual search as a core mechanism in multimodal {LLMs}.
\newblock In \emph{CVPR}, pages 13084--13094, 2024.

\bibitem[Xu et~al.(2025)Xu, Zhu, and Yang]{xu2025mcbench}
Yunqiu Xu, Linchao Zhu, and Yi Yang.
\newblock {MC-Bench}: A benchmark for multi-context visual grounding in the era of mllms.
\newblock In \emph{ICCV}, pages 17675--17687, 2025.

\bibitem[Yan et~al.(2026)Yan, Tong, Xue, Tang, Wang, Shi, Zhang, Li, and Zou]{yan2026metis}
Shilin Yan, Jintao Tong, Hongwei Xue, Xiaojun Tang, Yangyang Wang, Kunyu Shi, Guannan Zhang, Ruixuan Li, and Yixiong Zou.
\newblock Act wisely: Cultivating meta-cognitive tool use in agentic multimodal models.
\newblock \emph{arXiv preprint arXiv:2604.08545}, 2026.

\bibitem[Yang et~al.(2024)Yang, Kang, Huang, Xu, Feng, and Zhao]{yang2024depthanything}
Lihe Yang, Bingyi Kang, Zilong Huang, Xiaogang Xu, Jiashi Feng, and Hengshuang Zhao.
\newblock Depth anything: Unleashing the power of large-scale unlabeled data.
\newblock In \emph{CVPR}, pages 10371--10381, 2024.

\bibitem[Yang et~al.(2023)Yang, Song, Li, Zhao, Ge, Li, and Shan]{yang2023gpt4tools}
Rui Yang, Lin Song, Yanwei Li, Sijie Zhao, Yixiao Ge, Xiu Li, and Ying Shan.
\newblock {GPT4Tools}: Teaching large language model to use tools via self-instruction.
\newblock In \emph{NeurIPS}, 2023.

\bibitem[Yuan et~al.(2026)Yuan, Shen, Yan, Liu, Zhang, Fan, Xie, Wang, Guan, Wang, and Ye]{yuan2026dift}
Xiaosong Yuan, Chen Shen, Shaotian Yan, Kaiyuan Liu, Xiaofeng Zhang, Sinan Fan, Liang Xie, Wenxiao Wang, Renchu Guan, Ying Wang, and Jieping Ye.
\newblock Differential fine-tuning large language models towards better diverse reasoning abilities.
\newblock In \emph{ICLR}, 2026.

\bibitem[Zheng et~al.(2026)Zheng, Yang, Hong, Zhao, Xu, Yang, Shen, and Yu]{zheng2026deepeyes}
Ziwei Zheng, Minghao Yang, Jack Hong, Chenxiao Zhao, Guohai Xu, Le Yang, Chao Shen, and Xing Yu.
\newblock {DeepEyes}: Incentivizing ``thinking with images'' via reinforcement learning.
\newblock In \emph{ICLR}, 2026.

\bibitem[Zhou et~al.(2026)Zhou, Wei, Qu, Sakai, Kamigaito, and Watanabe]{zhou2026geopilot}
Xuan Zhou, Xuefeng Wei, Zhi Qu, Yusuke Sakai, Hidetaka Kamigaito, and Taro Watanabe.
\newblock A cross-domain tool-augmented vision--language framework for remote sensing image understanding.
\newblock \emph{Remote Sensing}, 18\penalty0 (10):\penalty0 1613, 2026.

\bibitem[Zuo et~al.(2025)Zuo, Kayan, Wang, Jeon, Deng, and Griffiths]{zuo2025towards}
Yiming Zuo, Karhan Kayan, Maggie Wang, Kevin Jeon, Jia Deng, and Thomas~L. Griffiths.
\newblock Towards foundation models for 3d vision: How close are we?
\newblock In \emph{International Conference on 3D Vision (3DV)}, pages 1285--1296, 2025.


\bibitem{bai2025qwen3vl}
Shuai Bai, Yuxuan Cai, Ruizhe Chen, Keqin Chen, Xionghui Chen, Zesen Cheng, Lianghao Deng, Wei Ding, et~al.
\newblock {Qwen3-VL} technical report.
\newblock \emph{arXiv preprint arXiv:2511.21631}, 2025.

\bibitem{lu2023mathvista}
Pan Lu, Hritik Bansal, Tony Xia, Jiacheng Liu, Chunyuan Li, Hannaneh Hajishirzi, Hao Cheng, Kai-Wei Chang, Michel Galley, and Jianfeng Gao.
\newblock {MathVista}: Evaluating mathematical reasoning of foundation models in visual contexts.
\newblock In \emph{ICLR}, 2024.

\bibitem{zhang2024mathverse}
Renrui Zhang, Dongzhi Jiang, Yichi Zhang, Haokun Lin, Ziyu Guo, Pengshuo Qiu, Aojun Zhou, Pan Lu, Kai-Wei Chang, Yu Qiao, Peng Gao, and Hongsheng Li.
\newblock {MathVerse}: Does your multi-modal LLM truly see the diagrams in visual math problems?
\newblock In \emph{ECCV}, pages 169--186, 2024.

\bibitem{qiao2024wemath}
Runqi Qiao, Qiuna Tan, Guanting Dong, Minhui Wu, Chong Sun, Xiaoshuai Song, Jiapeng Wang, Zhuoma GongQue, Shanglin Lei, YiFan Zhang, Zhe Wei, Miaoxuan Zhang, Runfeng Qiao, Xiao Zong, Yida Xu, Peiqing Yang, Zhimin Bao, Muxi Diao, Chen Li, and Honggang Zhang.
\newblock {We-Math}: Does your large multimodal model achieve human-like mathematical reasoning?
\newblock In \emph{Proceedings of the 63rd Annual Meeting of the Association for Computational Linguistics (Volume 1: Long Papers)}, pages 20023--20070, 2025.

\bibitem{zou2024dynamath}
Chengke Zou, Xingang Guo, Rui Yang, Junyu Zhang, Bin Hu, and Huan Zhang.
\newblock {DynaMath}: A dynamic visual benchmark for evaluating mathematical reasoning robustness of vision language models.
\newblock In \emph{ICLR}, 2025.

\bibitem{xiao2024logicvista}
Yijia Xiao, Edward Sun, Tianyu Liu, and Wei Wang.
\newblock {LogicVista}: Multimodal LLM logical reasoning benchmark in visual contexts.
\newblock \emph{arXiv preprint arXiv:2407.04973}, 2024.

\bibitem{qwen2026qwen35}
Qwen Team.
\newblock {Qwen3.5}: Towards native multimodal agents, 2026.
\newblock \url{https://qwen.ai/blog?id=qwen3.5}.

\end{thebibliography}
